\documentclass{article}
\usepackage{log_2023}	  

\usepackage{booktabs}            
\usepackage{multirow}            
\usepackage{amsfonts}            
\usepackage{graphicx}            

\usepackage{microtype}

\usepackage{multirow} 
\usepackage{caption}
\usepackage{subcaption}
\usepackage{float}

\usepackage{amsmath}
\usepackage{amssymb}
\usepackage{mathtools}
\usepackage{amsthm}

\usepackage{float}
\usepackage{xcolor}

\usepackage{algorithm}
\usepackage[noend]{algpseudocode}

\usepackage{tikz}
\usetikzlibrary{decorations.pathreplacing,calc,tikzmark}

\usepackage[
     backend=biber,
     style=numeric-comp,
     backref=true,
     natbib=true]{biblatex}
\title[Latent Space Representations of Neural Algorithmic Reasoners]{Latent Space Representations of Neural Algorithmic Reasoners}

\author[V. Mirjani\'c et al.]{%
Vladimir V. Mirjani\'c\\
University of Cambridge\\
\email{vvm22@cam.ac.uk}\And
Razvan Pascanu\\
Google DeepMind\\
\email{razp@google.com}\And
Petar Veli\v{c}kovi\'c\\
Google DeepMind\\
\email{petarv@google.com}
}

\newcommand*\Let[2]{\State #1 $\gets$ #2}

\newcommand{\entry}[2]{$#1\%$\tiny$\pm#2$}
\newcommand{\bestentry}[2]{${\mathbf{#1\%}}$\tiny$\pm#2$}
\newcommand{\uentry}[2]{$\mathcolor{blue}{#1\%}$\tiny$\pm#2$}
\theoremstyle{plain}
\newtheorem{theorem}{Theorem}[section]

\newtheorem{lemma}[theorem]{Lemma}

\theoremstyle{definition}

\theoremstyle{remark}

\begin{document}

\maketitle

\begin{abstract}
Neural Algorithmic Reasoning (NAR) is a research area focused on designing neural architectures that can reliably capture \emph{classical computation}, usually by learning to execute algorithms. A typical approach is to rely on Graph Neural Network (GNN) architectures, which encode inputs in high-dimensional latent spaces that are repeatedly transformed during the execution of the algorithm. In this work we perform a detailed analysis of the structure of the latent space induced by the GNN when executing algorithms. We identify two possible failure modes: (i) loss of resolution, making it hard to distinguish similar values; (ii) inability to deal with values outside the range observed during training. We propose to solve the first issue by relying on a \emph{softmax} aggregator, and propose to decay the latent space in order to deal with out-of-range values. We show that these changes lead to improvements on the majority of algorithms in the standard CLRS-30 benchmark when using the state-of-the-art Triplet-GMPNN processor.
\end{abstract}

\section{Introduction}

Algorithms are one of the cornerstones of computer science. Recently, a large body of work on Neural Algorithmic Reasoning showed how neural networks can be taught to simulate classical algorithms \cite{yan_neural_2020, velickovic_neural_2020, xu_what_2020}. This has emerged as a highly popular framework, and its utility extends beyond replicating steps of known algorithms. Neural networks trained on max-flow and min-cut have improved brain vessel classification \cite{numeroso_dual_2023}. In reinforcement learning, similar approaches are immune to many downsides of methods such as Value Iteration Networks \cite{deac_xlvin_2020}.

Neural Algorithmic Reasoners, often implemented as GNN architectures and trained to simulate algorithms, use latent spaces to represent complex data, with each of their message passing steps corresponding to a step in algorithm execution. The evolution of the latent representation therefore corresponds to the execution of an algorithm. In this work we exploit our understanding of classical algorithms in order to analyse this space.  Specifically, we use dimensionality reduction techniques like PCA and explore perturbations of algorithm inputs, to visualize and provide insights into the structures that emerge in the representational space. We observe clusters under different algorithmic symmetries, and we show that trajectories in the latent space tend to converge to a single attractor, using dynamical systems. 

Our analysis also reveals two weaknesses of the typical GNN architectures used for algorithmic reasoning. The first one is that these systems struggle when comparing similar values.
We hypothesize that due to the \emph{max} operator used for aggregation, the GNN tends to make a random choice between the values and is able to propagate gradients only through that choice. Even if the choice was suboptimal, it gets reinforced, and the learning process remains blind to the fact that it could perform better. To address this we propose to use \emph{softmax} as an aggregator, which in this scenario will allow gradient to propagate through all pathways, and the learning process to discover the optimal choice. 

The second weakness is that the model tends to struggle when encountering out-of-distribution values during algorithm execution. We propose that GNN should decay magnitude of the representations at each step, allowing slightly out-of-range values to become within range during the execution of the algorithm.  We show that these changes bring improvements to state-of-the-art models on the majority of algorithms from the commonly used CLRS-30 benchmark \cite{velickovic_clrs_2022, ibarz_generalist_2022}.

The success of our improvements indicates that a better understanding of latent spaces is likely to be crucial to further improving GNN architectures. In summary, our main contributions are
\begin{itemize}
    \item The first, comprehensive, study of the latent spaces emerging from training on algorithmic tasks, which shows that
    \begin{itemize}
        \item learned manifolds of graph embeddings are of much lower dimension compared to the size of latent spaces
        \item graphs with similar execution trajectories tend to cluster together
        \item the embeddings converge to an attractor state corresponding to the end of algorithm execution
    \end{itemize}
    \item Showing that GNNs fail to distinguish between branches of relatively similar values, and when encountering OOD values during execution.
    \item Softmax aggregation and processor decay as improvements, to ameliorate discovered failure modes, with an evaluation on CLRS-30 showing improvement on the majority of algorithms.
\end{itemize}

\section{Background and Related Work}

Consider the Bellman-Ford algorithm for finding shortest paths \cite{bellman_routing_1958}. Its input is a weighted graph with nodes $V$ and edges $E$, as well as a source node $s$. It outputs $\pi:V\to V$, pointers to predecessor nodes along shortest paths to source. Values $d:V\to\mathbb{R}$ denote the distances to source. The main update rule (see Algorithm~\ref{alg:bellman_ford}, lines 9-11) can be rewritten as in Eq.\ \ref{eq:bellman_update}.
\begin{equation}
    d_v = \min\left(\hat{d}_v,\ \min_{u\in\mathcal{N}_v}\left(\hat{d}_u + w(u,v)\right)\right)
    \label{eq:bellman_update}
\end{equation}

Here, $\hat{d}$ stands for the shortest distances from the previous step. Compare it to a general message passing GNN in Eq.\ \ref{eq:message_passing}, with $\psi$ and $\phi$ as trainable MLPs \cite{bronstein_geometric_2021}.

\begin{equation}
    \mathbf{z}^{(i+1)}_u=\phi\left(\mathbf{z}^{(i)}_u, \bigoplus_{v\in \mathcal{N}_u} \psi\left(\mathbf{z}_u^{(i)},\mathbf{z}^{(i)}_v\right)\right)
    \label{eq:message_passing}
\end{equation}

This similarity is no coincidence; GNNs are closely aligned with dynamic programming \cite{xu_what_2020, dudzik_graph_2022}. Therefore, they are particularly suited for simulating classical algorithms.

\cite{velickovic_neural_2020} uses this similarity to simulate algorithms by learning to replicate individual steps of their execution. For example, in Bellman-Ford (Algorithm~\ref{alg:bellman_ford}), one step corresponds to one iteration of the loop in lines 6-12, and the GNN is trained to reproduce intermediate values of $\pi$ and $d$. These intermediate values are called \emph{hints}. They guide a GNN in learning the algorithm, and they provide more immediate feedback than only using algorithm's outputs. In addition to hints, GNN is also given access to latent vectors in a high-dimensional space, to simulate working memory.

\begin{algorithm}[h]
  \caption{Bellman-Ford}
  \begin{algorithmic}[1]
    \Statex
    \Function{Bellman-Ford}{$\mathcal{G}(V, E)$, $s$}\tikzmark{right}
        \For {$v\in V$}
            \Let{$\pi_v$}{\textbf{None}}
            \Let{$d_v$}{$\infty$}
        \EndFor
        \Let{$d_s$}{0}
      \Repeat
        \Let{$\hat{d}$}{$d$}\tikzmark{POS1}
        \For{$e(u,v)\in E$}
            \If{$d_v > d_u + w(u,v)$}
                \Let{$d_v$}{$d_u + w(u,v)$}
                \Let{$\pi_v$}{$u$}
            \EndIf
        \EndFor \tikzmark{POS2}
      \Until{$\hat{d} = d$}\\
    \Return $\pi$
    \EndFunction
    \begin{tikzpicture}[overlay, remember picture]
        \node[anchor=base] (a) at (pic cs:POS1) {\vphantom{h}}; 
        \node[anchor=base] (b) at (pic cs:POS2) {\vphantom{g}}; 
        \draw [decoration={brace,amplitude=0.5em},decorate,ultra thick,blue]
         (a.north -| {pic cs:right}) -- (b.south -| {pic cs:right})
         node [align=center, text width=4.5cm, pos=0.5, anchor=west] {Step of algorithm execution};
    \end{tikzpicture}
  \end{algorithmic}
  \label{alg:bellman_ford}
\end{algorithm}

Several different approaches to aligning neural networks with algorithms exist. On a microscopic scale, there are Neural Execution Engines \cite{yan_neural_2020}, which learn basic operations, such as summation, $\min$, or $\max$, and combine them into larger compositional blocks. Macroscopic approaches also exist \cite{xu_what_2020}, and they learn algorithms end-to-end. The model we will focus on in this work, proposed by~\cite{velickovic_neural_2020}, that we will refer to as Neural Algorithmic Reasoner (NAR), operates on a mesoscopic scale, and learns key steps in algorithm execution.

These approaches target different levels of abstraction, and therefore their latent spaces learn to represent different types of data. When latent spaces of Neural Execution Engines are visualised with PCA, sequential natural numbers are found organised in chains in sorting tasks, and ``human-interpretable patterns'' are found for other tasks as well \cite{yan_neural_2020}. These patterns extend to values not seen during training, even with large holdout. Thus, these networks can structure their latent spaces even with low amounts of training data.

The reason we focus on the NAR is that it allows multi-task learning, while Neural Execution Engines are hard-wired for a specific algorithm. 
In fact, NAR architecture can learn all 30 algorithms from the CLRS-30 benchmark \cite{velickovic_clrs_2022} at once, and this even leads to improvement on some tasks, compared to learning them separately \cite{ibarz_generalist_2022}.
This suggests a much richer latent structure, which can provide insights that hold across multiple algorithms. 

\section{Latent Representations}
\label{sec:experiments}

We focus on a single architecture and a single algorithm.
For the algorithm, we pick Bellman-Ford (as implemented in Algorithm \ref{alg:bellman_ford}). Bellman-Ford trains quickly, since the number of algorithm steps is bounded by the number of nodes, and it uses little GPU memory while training. These properties make it a good choice for rapid prototyping of many different ideas.

For the network, we create a variation of PGN \cite{velickovic_pointer_2020}. We simplify the network by removing all ReLU non-linearities, so that the only nonlinearity remaining is the $\max$ aggregation inherent in the layer. We name the result LinearPGN. This new processor has simpler representations than original PGN, due to lack of non-linearities.

Despite being structurally simple, LinearPGN is expressive enough to simulate Bellman-Ford. Min aggregation over the edges in Bellman-Ford can be converted into max aggregation used in message passing by negating the messages both before and after it. The messages themselves are simply the distance updates $\hat{d}_u + w(u,v)$. This can be seen in Eq.\ \ref{eq:bellman_aligned}, which at the same time corresponds both to the Bellman-Ford update rule, and to fully linear message passing.

\begin{equation}
    \mathbf{z}_v^{(i+1)} = -\max_u\left(-\mathbf{z}_u^{(i)} - w(u,v)\right)
    \label{eq:bellman_aligned}
\end{equation}

\subsection{Trajectories of Embeddings}
\label{sec:trajs}

To visualise the latent spaces, we record the complete \emph{trajectories} of input graphs' latent embeddings, i.e., we save the hidden vectors $\mathbf{z}_v^{(i)}$ for each node $v$ in the graph, for each time step $i$. Some graphs may halt their execution earlier than others, and this skews the distribution of hidden vectors. To avoid this, we filter out a subset of graphs that all terminate at the same time $T$. Thus, trajectory data has shape $N\times |V|\times D\times T$, where $N$ is the number of sampled graphs, $|V|$ is the number of nodes in each graph, $D$ is the dimensionality of the latent space, and $T$ is the number of steps that the processor executed before terminating. We first reduce the number of dimensions by eliminating the (second) $|V|$ axis. We opt to use $\max$-aggregation, so that for each step $i$ we extract $\max_v \mathbf{z}_v^{(i)}$ out of the individual node embeddings. We provide an ablation to validate our choice in Appendix~\ref{sec:aggro}.

After this reduction, the trajectory tensor has shape $N\times D\times T$. We consider two different types of plot by

\begin{itemize}
    \item combining the dimension and time step axes; this leads to a matrix of size $N\times (D\cdot T)$ over which we can apply PCA. We call this the \textbf{trajectory-wise} plot because entire trajectories for each of $N$ input graphs are summarised into a single point. We reduce the number of $DT$ down to $d=2$ or $d=3$. 

    \item grouping time step and sample axes against the latent space, in order to obtain a matrix of size $(N\cdot T)\times D$. We call this the \textbf{step-wise} plot because each graph is represented with $T$ points -- one for each execution step. We typically trace red lines through the $T$ points of each graph, in order to represent trajectories in the latent space, and color the points based on the time step.
\end{itemize}

On all PCA visualisations, X axis is the axis of the highest variance, and Y axis has the second highest variance.

\begin{figure*}[h!]
    \centering

    \includegraphics[width=\textwidth]{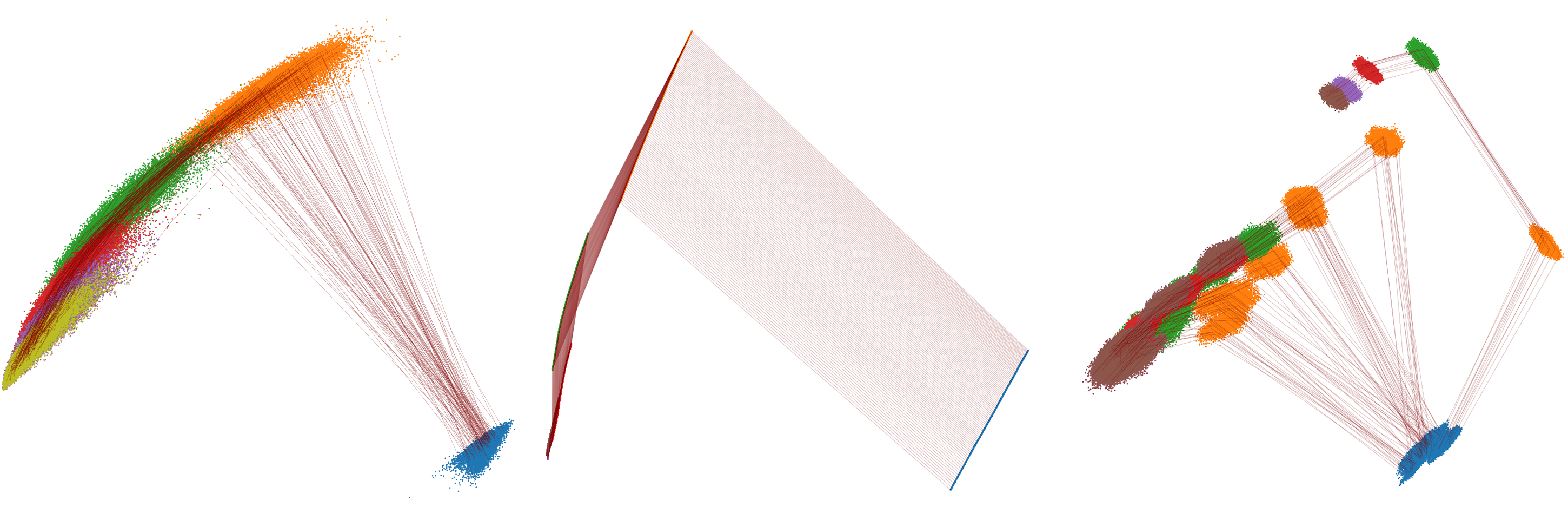}
     
    \caption{Left: Step-wise PCA visualisation of latent spaces for random graphs. Middle: Step-wise PCA for a set of graphs derived from scaling symmetry. Scaling has one degree of freedom. This is captured in the latent space, as all step embeddings are one-dimensional. Right: Step-wise PCA of 8 classes of random graphs, each containing graphs equivalent under reweighting symmetry. Clusters corresponding to each class of graphs are clearly visible, showing that the model learns to group graphs with same execution close together. Red lines trace graph trajectories. Different colors correspond to different steps in algorithm execution. Sequential view available in Appendix \ref{sec:sideways}.}
    \label{fig:1}
\end{figure*}

\subsection{Dimensionality}

We first study the robustness of latent representations. The latent space in NAR is $128$-dimensional ($D=128$), but valid trajectories might live in a considerably lower-dimensional space, where the redundancy might help robustness to noise. 

We measure the amount of variance captured by three most dominant dimensions for trajectory-wise and step-wise PCA. We sample random Erd\H os-R\'enyi graphs, and measure that in trajectory-wise PCA the three dimensions capture $63.5\%$ of variance. For step-wise PCA (Fig.\ \ref{fig:1} left), this number is $96.4\%$. If the embeddings were uniformly spread, then each PCA dimension would account for only $\frac{1}{128}$ of the total variance, and therefore the first three dimensions would contribute with only $2.3\%$. The $63.5\%$ that they contribute with is therefore quite high, and indicates that the representations are low-dimensional. The even higher value of over $96.4\%$ for step-wise PCA shows that there is a large amount of structure in the step-wise view of the latent space, and that the movement between steps is significantly stronger than the variance in graph trajectories.

\subsection{Representations Under Symmetries}

GNNs are designed to be invariant or equivariant to graph permutations. Therefore, they can operate on all types of graphs, including algorithm inputs. However, algorithms and their data often have more structure than mere permutation invariance. After showing the structure of latent spaces for random graphs (Fig.\ \ref{fig:1} left), we explore how the model deals with additional symmetries of the learned algorithm.
Recall that inputs to Bellman-Ford are weighted graphs, with graph weights in the interval $(0,1)$. There are many non-trivial transforms that we can perform on the inputs that keep algorithm execution constant, but change the embeddings in the latent space.

We start by formalizing the concept of \textit{keeping algorithm execution constant}. We take it to mean that if we apply a transformation to the inputs, then hints at each time step should change accordingly. Bellman-Ford has two kinds of hints, $\pi: V\to V$, pointers to other nodes, and $d: V\to\mathbb{R}$, distances to source. The difference between them impacts how they change with changes in inputs.
By analogy with the permutation equivariance of GNNs, we require that both $\pi$ and $d$ be equivariant to permutations of input nodes. Meanwhile, $\pi$ should be invariant and $d$ equivariant to transformations that change weights only. To understand how symmetries are encoded in the latent space, we study collections of inputs that have equivalent algorithm executions. We skip over permutation symmetries because they are general to all GNNs, and look at those symmetries specific to Bellman-Ford.

\subsubsection{Scaling Symmetry}
\label{sec:scaling_sym}

First we consider the scaling symmetry of graph weights $w_{uv}\mapsto\lambda w_{uv}$, where $\lambda>0$, which trivially preserves algorithm execution. As discussed above, this symmetry preserves pointers $\pi$, while transforming distances $d$ as $d_u\mapsto\lambda d_u$. To visualise the symmetry (Fig.\ \ref{fig:1} middle), we generate one random graph, and then scale it with $\lambda\in(\frac{1}{2},1)$. We avoid $\lambda>1$ as that would create edge weights larger than those seen in training. We also avoid small $\lambda$ because $w=0$ is used to encode lack of edges, and not edges of weight $0$. This means that representations around $w=0$ can behave unpredictably.

Note that scaling by positive $\lambda$ is necessary because introducing negative values creates negative cycles that dramatically change algorithm execution. Also, observe that adding a constant $c$ to each edge is not an algorithm symmetry. A single degree of freedom of this symmetry, $\lambda$, gives rise to one-dimensional embeddings (Fig.\ \ref{fig:1} middle). The embeddings are not in a straight line, but rather on a non-trivial curve. This is not as visible in the step-wise plot, as movement between steps overshadows variance between graphs.

\subsubsection{Reweighting Symmetry}

Next we turn to a more complex symmetry of Bellman-Ford. We name it reweighting symmetry after the ``reweighting'' trick from the Johnson's algorithm \cite{johnson_efficient_1977}. We outline the trick in Lemma \ref{thm:reparam}.   

\begin{lemma} 
\label{thm:reparam}
    (Reweighting trick)  Let $h: V\to \mathbb{R}$ assign arbitrary values to nodes. Suppose the input graph $\mathcal{G}$ with source node $s$ is reweighted as follows
    
    $$\hat{w}(u,v)\triangleq w(u,v) + h(u)-h(v)$$
    
    Then the new re-weighted graph $\mathcal{\hat{G}}$ has identical execution trace, meaning that at each execution step $\hat\pi=\pi$ and $\hat{d}(u)=d(u)+h(s)-h(u)$ hold, i.e. while values change, branches taken by the algorithm do not.
\end{lemma}

\begin{figure*}
    \centering

    \includegraphics[width=\textwidth]{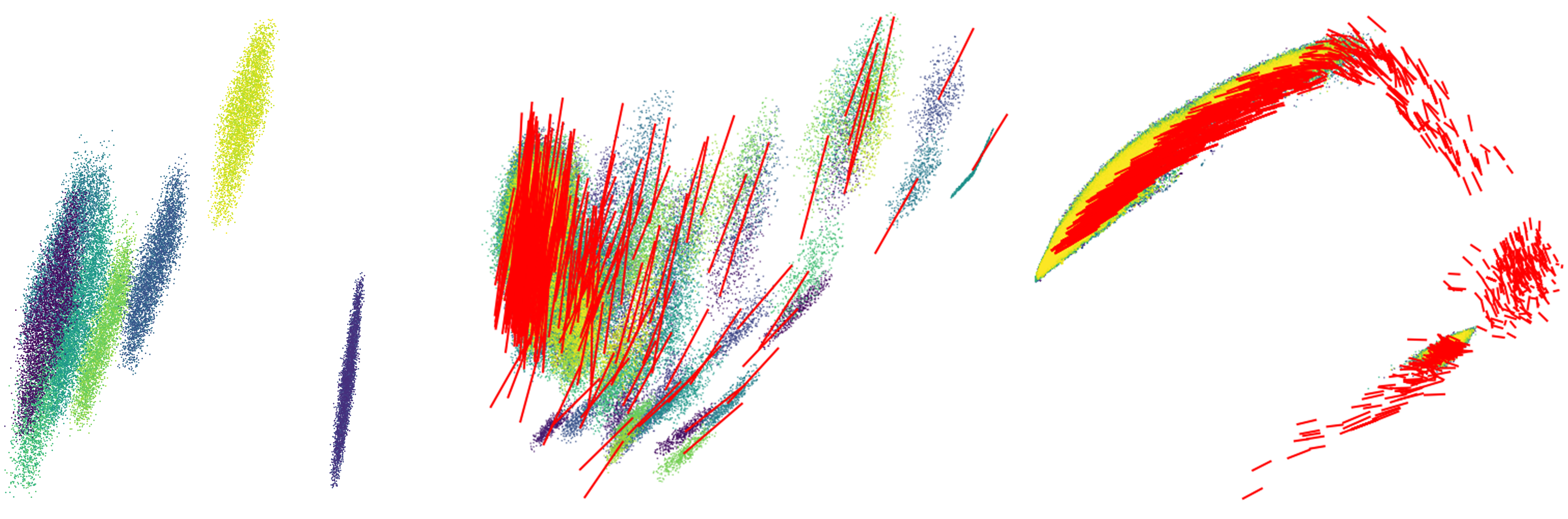}
     
    \caption{Left: Trajectory-wise PCA of eight clusters of reweighted graphs showing that they all contain a single dominant direction. Different clusters have different colors. Middle: Many embedding clusters with dominant directions overlaid in red. Right: Step-wise PCA of random graphs with the dominant cluster directions overlaid in red.}
    \label{fig:2}
\end{figure*}

In Johnson's algorithm, this trick removes negative edges from the graph, so that Dijkstra's algorithm can be used to find shortest paths. We use the trick to generate many graphs with the same execution.

The reweighting trick operates on graphs with real-valued weights, but our Bellman-Ford NAR is trained on graphs with weights in $(0,1)$. Therefore, we must ensure that weights stay in this distribution when applying Lemma \ref{thm:reparam} to generate perturbations. That is why we sample weights in the reduced range $(c, 1-c)$, and then sample $h$ randomly from $(0, c)$. Thus, we preserve high variance between weights from the original graph and can safely apply reweighting, as we are guaranteed to stay in the $(0,1)$ range afterwards. We experiment with $c=\frac{1}{2}$ and $c=\frac{1}{4}$, and observe no difference.

We sample several such random graphs and then generate multiple variations through reweighting (Fig.\ \ref{fig:1} right). These reweighting symmetries have $|V|$ degrees of freedom because each node $v$ gets assigned an independent value $h(v)$. In line with that, variance of $94\%$ for the three most dominant dimensions of reweighted graphs is similar to that of random graphs. However, it is highly significant that these clusters are clearly separated.

\subsection{Cluster Directions in Latent Spaces}

Another interesting observation to be made regarding reweighted graphs is that their embeddings seem to have one main direction of variance. This is more easily observed in the trajectory-wise plot (Fig.\ \ref{fig:2} left) than in the step-wise plot. Note that the variance along these directions is larger than the difference between cluster embeddings. We are interested in learning whether these directions are parallel or perpendicular to the manifold describing the execution of the algorithm. If they are perpendicular, moving along them should not affect the execution of the algorithm and hence the behavior of the model. 

To proceed, we build a database of hundreds of reweighting clusters in order to cover as much of the embeddings manifold as possible. Then, we apply PCA to each cluster at each time step, and extract the first several principal components. We use this fine-grained approach because the manifold itself is curved, and therefore these directions vary along it (Fig.\ \ref{fig:2} right). Qualitatively, these principal directions seem to be parallel to the manifold, which would mean that moving along them changes the meaning of the embeddings. Nevertheless, we perform a quantitative analysis using our database of directions.

We modify the embeddings by adding noise at test-time along these principal directions versus along random directions. We also consider removing the variance along the directions of interest, and also preserving only the variance along these directions. To pick the correct direction representing the symmetry, we pick the principal direction of the closest cluster from the database. In a control experiment, we pick a mean of all directions instead.

\begin{table*}[htbp]
  \centering
  \caption{Accuracy with perturbations along principal directions of variance. Perturbations include: the noise-free baseline (Noise-free), Gaussian noise along the direction (Directional), Gaussian noise along random direction (Random), projecting out the direction (Out), and projecting the embedding onto the direction (Onto).}
    \vspace{0.5em}
    \begin{tabular}{lccccc}
        \toprule
        \multicolumn{1}{c}{} & \multicolumn{5}{c}{Perturbation Type}\\
        \cmidrule(rl){2-6} 
         Direction  & {Noise-free} & {Directional} & {Random} & {Out} & {Onto} \\
        \cmidrule(r){1-1} \cmidrule(rl){2-6}
          L2 Closest & \multirow{2}{*}{\entry{93.58}{3.41}} & \entry{88.55}{6.28} & \multirow{2}{*}{\entry{93.17}{3.67}} & \entry{63.46}{7.10} & \entry{71.74}{6.91} \\
          Mean & & \entry{90.60}{5.16} & & \entry{5.20}{4.47} & \entry{26.07}{5.91} \\
        \bottomrule
    \end{tabular}
  \label{tab:1}
\end{table*}

Adding noise along these directions is more harmful than adding noise along a random direction. Also, projecting embeddings onto the directions is better than projecting the directions out of them (Tab.\ \ref{tab:1}). Both of these results contradict the hypothesis that these directions might be perpendicular to the manifold. One interpretation could be that the manifold is highly curved, or that we can view it as divided into small patches. Any movement within the path preserves execution, while larger movements reach different patches with different algorithm trajectories. 

\section{Mispredict Analysis}

We now investigate how close the model learns to be aligned with the ground-truth Bellman-Ford algorithm. To do so, we consider three faulty implementations of the Bellman-Ford algorithm, all of which represent simplifications that the model might have learned, and mostly change how distances are being compared (Alg.~\ref{alg:bellman_ford} line 9). Firstly, we discuss the Greedy-Ford algorithm, which compares only weights and not full distances, and is used to see whether the model learns a simpler algorithm than what is required. Secondly, the Decay-Ford algorithm scales distances by a constant, and is used to check whether the model learns the correct algorithm while falsely learning a few weights. Finally, the Noisy-Ford algorithm, which adds small amount of noise to distances, is there to check whether errors of the model result from noise in the embeddings.

\begin{figure}[htbp]
    \centering
    \includegraphics[width=\textwidth]{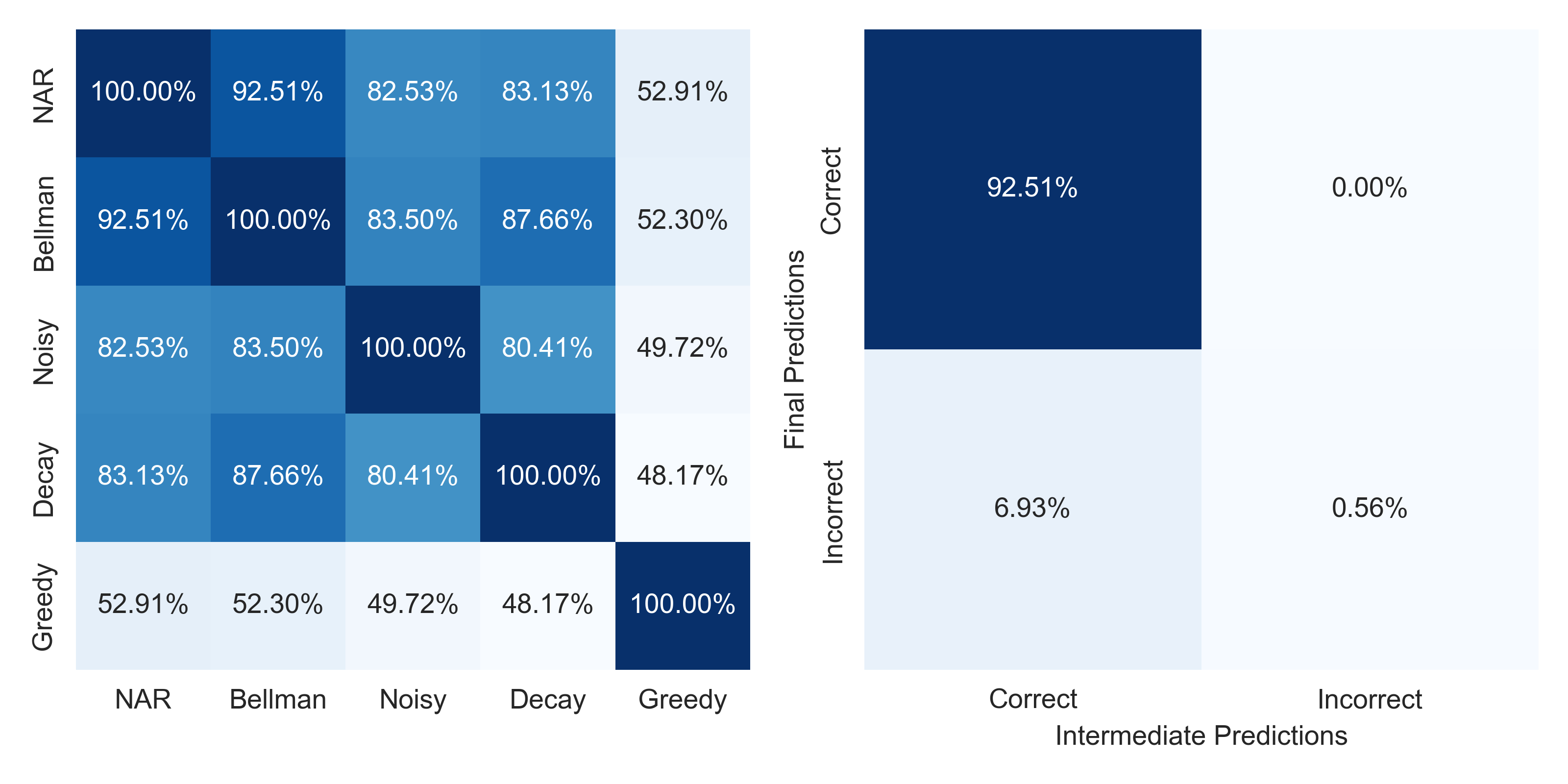}
    \caption{Left: Similarity between final predictions for our model, true Bellman-Ford, and multiple faulty implementations. Our model is closer to ground truth than to any others. Right: Predictions of the model split on the basis of whether they match ground truth at the end (y-axis), and during execution (x-axis). Almost all mispredicts were correctly predicted at an earlier point. }
    \label{fig:4}
\end{figure}

Our model is close in its predictions to all variants except Greedy-Ford, which means that it is learning something quite similar to the true algorithm (Fig.\ \ref{fig:4} left). In fact, out of all variants we have considered, our model is the closest to ground truth.
After this, we separate predictions based on whether they are correct at the end of algorithm execution, and whether they are correct at any intermediate step (Fig.\ \ref{fig:4} right). Intermediate predictions also include final predictions, which means that the upper-right field is exactly $0\%$ (if our model's prediction for a node is never correct, then it is also not correct in the end). However, the bottom-left field is surprisingly high.

This shows that nearly all mispredicts were predicted correctly at some point during the simulation, and were then changed to a wrong value afterwards. Only one percent of nodes were never predicted correctly. The ground-truth Bellman-Ford algorithm, meanwhile, never changes the correct output.

We observe that these wrong predictions typically occur when the network has to decide between two different paths with similar distances, and give an example in Appendix \ref{sec:example}. To address this problem, we propose as a solution to use \textbf{softmax aggregation} instead of hard $\max$. With $\max$ aggregation, derivatives are backpropagated only through the largest value, even if others are very close to it, and information can be lost. With softmax, all values impact the output and all values are backpropagated through. The effect would be particularly strong in cases where there are two large values and many small ones. We regard this as one main outcome of our analysis and validate it at scale in Section~\ref{sec:eval}.

\subsection{Mispredicts and Value Generalisation}

Consider random weighted graphs where each pair of nodes is connected with probability $p$, such as those used in our evaluations. Let weights be sampled uniformly from $(0,1)$, and $p=0.5$. We measure the average distance between nodes to be $0.15$. However, if $p=0.25$, the average distance rises to $1.1$. 

This shift in graph statistics means that a model trained on random graphs where $p_{avg}=0.5$ does not see long distances as often as short ones and struggles when handling them. Indeed, as we change connectivity of graphs from $p=0.5$ to $p=0.25$ during testing, we notice a drop in accuracy from $93\%$ to $87\%$. On the other hand, if we also scale weights down before evaluating, to mitigate the distance increase, the accuracy is restored to previous levels. This shows that the drop in performance is due to the model failing to handle paths longer than those it has seen during training.

\begin{figure}[htbp]
    \centering
    \includegraphics[width=\textwidth]{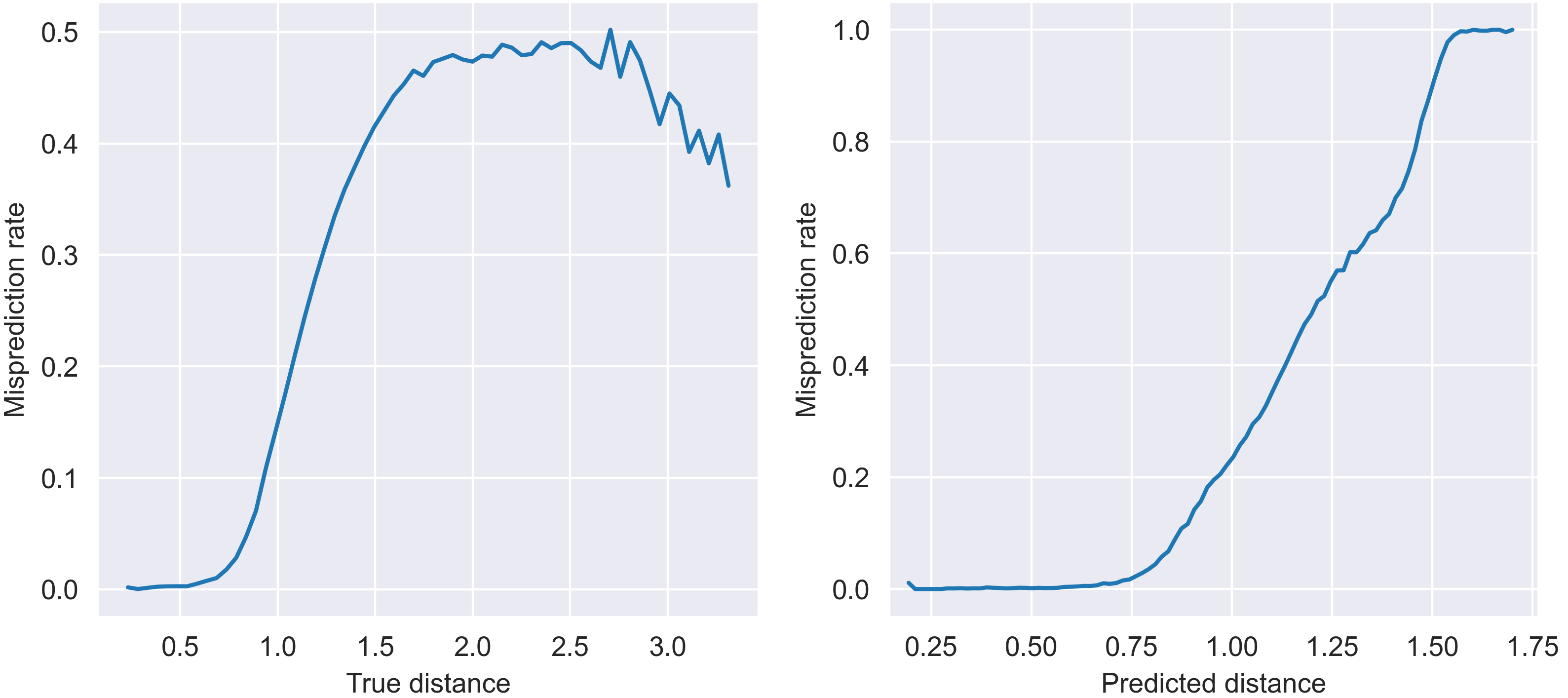}
    \caption{Left: Misprediction rate as a function of the shortest path distances. Its theoretical range is from $0$ to $1$, and lower is better. Right: Misprediction rate as a function of the \emph{predicted} distances.}
    \label{fig:5}
\end{figure}

In fact, wrong predictions of the model on Bellman-Ford algorithm with LinearPGN as the processor are directly correlated with the nodes' distances to source (Fig.\ \ref{fig:5}). We observe that the model is failing to correctly reproduce distances larger than $1$. Furthermore, we show that misprediction rate steadily increases for predicted distances between $0.75$ and $1.5$, and for paths with predicted distance above $1.5$ the model wrongly predicts every single one (Fig.\ \ref{fig:5} right).

As the model is struggling with large out-of-distribution values, we propose using a decay-like regularisation where, at every message passing step, we scale the embeddings by a constant $c<1$. We show in the following section that this provides improvements not only on the Bellman-Ford algorithm, but on other algorithms in the CLRS-30 benchmark \cite{velickovic_clrs_2022} as well.

\section{Evaluation on CLRS-30}
\label{sec:eval}

Our investigation produced two proposed changes that have the potential to improve the ability of GNN to model algorithms.

The first is softmax aggregation. In GNNs, once all messages to node $v_i$ are computed, they are aggregated in order to produce the new latent embeddings. Common choices for aggregation include summing over the messages, or finding the mean, but $\max$ aggregation is most commonly used in algorithmic reasoning tasks. That is, latents are calculated via $\mathbf{z}_i = \max_{j} \mathbf{m}_{ji}$. The way messages $\mathbf{m}$ are computed depends on the layer (PGN \cite{velickovic_pointer_2020}, Triplet-GMPNN \cite{ibarz_generalist_2022}, etc.). We argue that $\max$ aggregation cuts off messages that are close but not equal to the maximum value, and that this explains why the model struggles when comparing paths of similar lengths in the Bellman-Ford algorithm. 

Softmax aggregation \cite{li_deepergcn_2020} weighs each message based on its softmax coefficient and then sums them up:

\begin{equation}
    z_{i} = \sum_j \sigma(\frac{\mathbf{m}_{-i}}{T}, j) \cdot \mathbf{m}_{ji}
    \label{eq:softmax}
\end{equation}

We write softmax of $x_1$, $x_2$, ..., as $\sigma(x_{-}, i) = \frac{\exp{x_i}}{\sum_j \exp{x_j}}$. This is further parameterised with temperature $T$. As $T\to 0$, softmax approaches the values of (hard) $\max$, while for $T=1$ it behaves as standard unparametrised softmax, and as $T\to\infty$ it behaves as a mean aggregator.

The second improvement we proposed is the use of processor decay. This means that after each reduction, we scale the values down with a constant factor $c$.

These modifications can be applied individually, or together. For evaluating them, we use the CLRS-30 benchmark \cite{velickovic_clrs_2022}, which contains a curated list of 30 classical algorithms \cite{cormen_introduction_2009}. These algorithms cover a wide range of concepts -- searching, sorting, dynamical programming, divide and conquer, greedy, as well as many graph-specific topics, such as shortest paths, spanning trees, and more. All algorithms are tested on graphs that are four times larger than those seen during training. We estimate standard deviation with five runs. We use Triplet-GMPNN \cite{ibarz_generalist_2022} as the NAR processor for evaluation, as it is currently state-of-the-art on CLRS-30.

\begin{table}[htbp]
  \centering
  \caption{Performance of Triplet-GMPNN with pipeline modifications on CLRS-30 algorithms. Pipeline modifications are softmax aggregation (sft) and processor decay (dec).}
    \vspace{0.5em}
    \begin{tabular}{lcccc}
        \toprule
         Summary       & Baseline & +sft & +dec & +sft+dec \\
        \cmidrule(rl){1-1} \cmidrule(rl){2-5}
        Best on  \#/30 & 11 & 8 & 2 & 9 \\
        Above baseline & N/A & 15 & 10 & 11 \\
        \bottomrule
    \end{tabular}
  \label{tab:2}
\end{table}

We observe that our changes achieve top accuracy on the majority of tasks (Tab.\ \ref{tab:2}). Furthermore, softmax is most advantageous when used alone. Decay is more situational, and more work is needed in order to understand how to separate its benefits from the downsides. We use no hyperparameter tuning for training. Softmax temperature is set to $0.01$, and processor is decayed with a factor of $0.9$.

Finally, we perform multiple ablations. We show that small decay and small temperature perform better than large ones. We also show that our analysis of latent spaces extends to Triplet-GMPNN. Full results can be found in the Appendix.

\section{Discussion}

\textbf{Conclusions}\quad We provide an extensive analysis of the latent representations learned by GNN trained on algorithmic reasoning tasks. We show that these latent spaces have rich structure, in particular in how they reflect different symmetries in executing algorithms. We show that the execution trajectories in the latent space tend to converge to an attractor-like state, and graphs of similar executions tend to cluster together. Furthermore, 
from our analysis we identify two issues with current architectures: (i) dealing with values of similar magnitude, and (ii) dealing with values that are out of range compared to those seen during training. 

We hypothesise that the first issue is due to the use of the max aggregator function, which backpropagates gradients only along the largest of the similar values, making it harder for the learning process to identify whether it made a suboptimal choice. We propose to use \emph{softmax} instead of \emph{max} as aggregator, allowing gradients to flow through all paths proportional to their magnitude. We empirically validate this choice at scale on the CLRS-30 benchmark. 

The second issue for the Bellman-Ford algorithm happens when accumulating distances between nodes. The issue is that depending on the graph connectivity the distribution of distances and the embeddings in latent space can change drastically. We propose a simple fix -- decaying the magnitude of the embedding by a fixed rate at every execution step. This allows the embeddings to consistently stay in a similar range. Decay, coupled with the choice of softmax aggregator, provides improvement across many algorithms in CLRS-30 benchmark.

We regard our analysis as the first step in trying to understand how neural architectures model the algorithms they are trained to mimic. Our work shows that such analysis can reveal weaknesses of current parametrisations, and it points towards potential ways of addressing them. 

\textbf{Future Work}\quad We demonstrated that the latent space contains a single attractor to which the trajectories converge. This convergence could be safely exploited as a termination criterion. Currently, all experiments using CLRS-30 rely on knowing the exact number of steps needed for each input, even at test time. This is unrealistic, as such information is not available in the real world. An alternative, explored in \cite{velickovic_neural_2020}, is to use a dedicated neural network to learn when to terminate. The authors proposed using embeddings after each step as an input to the network. Our research, meanwhile, motivates the use of differences between successive encodings. Another direction for future work is investigating the impact of encoder and decoder on our analysis. In this project, we mainly focused on the processor, assuming that it learns the bulk of the algorithm. However, the importance of encoders and decoders could also be investigated (e.g. by freezing them during training). Finally, this study can be extended to multi-task learning.

\section*{Acknowledgements}

We would like to thank Prof.~Pietro Li\`o (University of Cambridge) for supervising the project while it was being done in partial fulfillment of the requirements for the Computer Science Tripos, Part III (MEng). We would like to thank Dr.~Jovana Mitrovi\'c and Prof.~Karl Tuyls (Google DeepMind) for conducting the tech and strategy reviews, and providing useful comments.


\nocite{bellman_routing_1958, bishop_pattern_2006, bronstein_geometric_2021, cormen_introduction_2009, dudzik_graph_2022, goodfellow_deep_2016, johnson_efficient_1977, maaten_visualizing_2008}

\printbibliography

\newpage
\appendix
\section{Detailed Results on CLRS-30}

Here we evaluate softmax aggregation and processor decay on Triplet-GMPNN and MPNN architectures. Best results are marked in \textbf{bold}, while (non-best) results above baseline are \textcolor{blue}{blue}.

\subsection{Results on Triplet-GMPNN}

\begin{table}[H]
  \centering
  \caption{Performance of Triplet-GMPNN with pipeline modifications on CLRS-30 algorithms. Pipeline modifications are softmax aggregation (sft) and processor decay (dec). Standard deviation estimated with five runs. We observe minor differences between the baseline accuracy as measured by us and that of \cite{ibarz_generalist_2022}. We expect that the cause lies in \cite{ibarz_generalist_2022} performing 10 runs as opposed to our five runs.}
    \vspace{0.5em}
    \begin{tabular}{lcccc}
        \toprule
        \multicolumn{1}{c}{} & \multicolumn{4}{c}{Triplet-GMPNN}\\
        \cmidrule(rl){2-5} 
         Algorithm             & baseline & +sft & +dec & +sft+dec \\
        \cmidrule(r){1-1} \cmidrule(rl){2-5}
Activity Selector & \bestentry{96.46}{1.64} & \entry{95.48}{1.17} & \entry{88.83}{4.37} & \entry{90.50}{3.67} \\
Articulation Points & \entry{68.21}{33.58} & \bestentry{73.10}{21.77} & \uentry{68.28}{7.89} & \entry{59.99}{8.26} \\
Bellman-Ford & \entry{98.68}{0.34} & \uentry{98.80}{0.14} & \entry{84.53}{31.51} & \bestentry{98.96}{0.35} \\
BFS & \entry{99.46}{0.41} & \bestentry{99.63}{0.10} & \uentry{99.54}{0.16} & \uentry{99.50}{0.09} \\
Binary Search & \entry{60.98}{7.91} & \uentry{64.64}{8.75} & \entry{58.88}{19.43} & \bestentry{66.49}{6.29} \\
Bridges & \entry{74.31}{17.35} & \uentry{77.83}{14.64} & \uentry{74.50}{30.95} & \bestentry{84.10}{18.15} \\
Bubble Sort & \bestentry{66.34}{6.29} & \entry{56.15}{10.39} & \entry{46.65}{29.37} & \entry{55.51}{17.15} \\
DAG Shortest Paths & \bestentry{89.17}{6.20} & \entry{83.18}{7.46} & \entry{77.13}{15.14} & \entry{82.56}{7.53} \\
DFS & \bestentry{19.29}{7.94} & \entry{12.88}{3.44} & \entry{15.68}{5.96} & \entry{20.65}{8.01} \\
Dijkstra & \entry{96.51}{1.68} & \bestentry{97.99}{0.24} & \entry{94.46}{4.16} & \entry{96.02}{2.03} \\
Find Maximum Subarray & \bestentry{62.84}{6.45} & \entry{58.55}{5.99} & \entry{48.89}{5.83} & \entry{52.77}{3.30} \\
Floyd-Warshall & \entry{8.55}{4.82} & \uentry{25.77}{16.82} & \uentry{18.90}{9.45} & \bestentry{32.52}{13.77} \\
Graham Scan & \entry{91.89}{2.57} & \bestentry{93.39}{3.04} & \entry{88.91}{2.55} & \entry{87.21}{7.70} \\
Heapsort & \bestentry{42.80}{18.85} & \entry{40.19}{21.86} & \entry{28.07}{25.11} & \entry{16.28}{11.72} \\
Insertion Sort & \bestentry{69.42}{7.87} & \entry{61.17}{17.13} & \entry{54.75}{22.40} & \entry{61.69}{9.92} \\
Jarvis March & \entry{90.58}{1.79} & \bestentry{90.86}{1.73} & \entry{64.17}{30.05} & \entry{81.57}{13.63} \\
KMP Matcher & \entry{10.70}{7.69} & \entry{4.16}{2.40} & \uentry{11.20}{9.77} & \bestentry{20.59}{11.94} \\
LCS Length & \entry{87.20}{0.84} & \uentry{87.41}{0.92} & \entry{87.11}{1.71} & \bestentry{87.75}{0.97} \\
Matrix Chain Order & \entry{92.45}{1.11} & \entry{91.81}{1.07} & \bestentry{92.90}{1.49} & \uentry{90.45}{2.77} \\
Minimum & \entry{93.29}{4.16} & \uentry{95.77}{1.06} & \uentry{95.46}{5.78} & \bestentry{96.08}{2.07} \\
MST Kruskal & \bestentry{88.83}{2.29} & \entry{87.96}{0.73} & \entry{87.00}{2.32} & \entry{85.49}{2.63} \\
MST Prim & \entry{86.41}{4.43} & \bestentry{87.42}{2.02} & \entry{79.43}{6.33} & \entry{76.68}{16.49} \\
Na\"ive String Matcher & \bestentry{17.50}{12.11} & \entry{8.52}{6.56} & \entry{9.85}{3.19} & \entry{10.16}{6.94} \\
Optimal BST & \bestentry{82.93}{1.84} & \entry{82.90}{0.85} & \entry{82.52}{1.59} & \entry{81.82}{2.21} \\
Quickselect & \entry{0.78}{0.94} & \entry{0.63}{0.82} & \uentry{1.37}{2.14} & \bestentry{1.62}{1.21} \\
Quicksort & \entry{43.54}{10.82} & \bestentry{49.22}{13.53} & \entry{31.76}{18.68} & \uentry{46.46}{24.85} \\
Segments Intersect & \entry{98.59}{0.22} & \uentry{98.74}{0.22} & \uentry{98.68}{0.29} & \bestentry{98.81}{0.15} \\
Strongly Connected & \entry{43.23}{6.83} & \entry{37.26}{9.81} & \bestentry{43.63}{7.94} & \entry{32.53}{5.03} \\
Task Scheduling & \entry{87.26}{3.39} & \bestentry{87.58}{0.66} & \entry{86.88}{1.46} & \entry{86.34}{1.24} \\
Topological Sort & \bestentry{71.29}{3.25} & \entry{71.25}{7.05} & \entry{69.38}{8.40} & \entry{64.27}{6.15} \\
        \bottomrule
    \end{tabular}
  \label{tab:results}
\end{table}

\subsection{Results on MPNN}

On Table \ref{tab:results_mpnn} we display our results for MPNN instead of Triplet-GMPNN. Generally speaking, improvements are more significant compared to Triplet-GMPNN.

\begin{table}
  \centering
  \caption{Performance of MPNN with pipeline modifications on CLRS-30 algorithms. Pipeline modifications are softmax aggregation (sft) and processor decay (dec). Standard deviation estimated with five runs.}
    \vspace{0.5em}
    \begin{tabular}{lcccc}
        \toprule
        \multicolumn{1}{c}{} & \multicolumn{4}{c}{MPNN}\\
        \cmidrule(rl){2-5} 
         Algorithm             & baseline & +sft & +dec & +sft+dec \\
        \cmidrule(r){1-1} \cmidrule(rl){2-5}
Activity Selector & \entry{88.57}{5.13} & \uentry{90.17}{4.49} & \bestentry{92.10}{1.59} & \uentry{91.56}{2.04} \\
Articulation Points & \entry{38.82}{19.85} & \uentry{46.94}{7.27} & \entry{30.55}{18.38} & \bestentry{47.37}{13.24} \\
Bellman-Ford & \entry{92.49}{0.99} & \bestentry{93.37}{1.33} & \uentry{92.76}{0.72} & \uentry{92.91}{1.04} \\
BFS & \entry{99.56}{0.17} & \uentry{99.71}{0.10} & \uentry{99.69}{0.16} & \bestentry{99.74}{0.14} \\
Binary Search & \entry{46.89}{10.75} & \uentry{48.85}{7.60} & \uentry{54.30}{8.30} & \bestentry{56.89}{9.37} \\
Bridges & \entry{23.31}{13.09} & \uentry{23.54}{13.19} & \bestentry{24.02}{13.43} & \uentry{23.83}{13.34} \\
Bubble Sort & \entry{64.59}{3.63} & \uentry{65.55}{12.71} & \bestentry{66.82}{6.76} & \entry{60.34}{17.74} \\
DAG Shortest Paths & \entry{81.55}{10.00} & \bestentry{92.01}{0.95} & \uentry{85.42}{7.51} & \uentry{84.05}{3.39} \\
DFS & \entry{13.77}{3.43} & \uentry{14.66}{7.05} & \uentry{16.05}{2.53} & \bestentry{16.64}{6.04} \\
Dijkstra & \bestentry{93.93}{1.41} & \entry{93.44}{2.34} & \entry{93.48}{3.08} & \entry{92.54}{1.64} \\
Find Maximum Subarray & \bestentry{51.66}{7.19} & \entry{50.35}{7.55} & \entry{50.09}{1.67} & \entry{47.44}{5.54} \\
Floyd-Warshall & \entry{15.25}{6.47} & \uentry{16.74}{11.31} & \uentry{18.74}{7.18} & \bestentry{19.37}{4.00} \\
Graham Scan & \entry{92.62}{1.35} & \bestentry{93.87}{1.22} & \entry{89.23}{7.46} & \uentry{93.13}{2.28} \\
Heapsort & \entry{66.02}{3.51} & \uentry{66.51}{4.48} & \bestentry{68.87}{6.86} & \entry{62.26}{4.57} \\
Insertion Sort & \entry{55.96}{14.11} & \entry{47.48}{22.99} & \uentry{60.50}{7.06} & \bestentry{66.64}{5.33} \\
Jarvis March & \bestentry{95.29}{1.37} & \entry{94.21}{1.83} & \entry{95.05}{1.28} & \entry{94.89}{0.95} \\
KMP Matcher & \bestentry{8.83}{3.11} & \entry{5.94}{1.42} & \entry{7.52}{4.01} & \entry{6.41}{3.26} \\
LCS Length & \entry{80.05}{5.02} & \entry{79.31}{6.58} & \bestentry{84.76}{2.45} & \entry{79.12}{5.73} \\
Matrix Chain Order & \entry{84.17}{0.79} & \bestentry{84.80}{1.37} & \entry{83.56}{2.14} & \uentry{84.48}{0.95} \\
Minimum & \entry{94.31}{5.13} & \bestentry{95.62}{0.15} & \entry{93.74}{5.05} & \entry{93.49}{3.54} \\
MST Kruskal & \entry{63.30}{23.57} & \uentry{66.65}{11.25} & \bestentry{68.41}{8.95} & \entry{61.04}{22.13} \\
MST Prim & \bestentry{72.39}{11.88} & \entry{67.10}{10.57} & \entry{61.05}{13.18} & \entry{65.82}{15.66} \\
Na\"ive String Matcher & \entry{5.97}{3.57} & \entry{5.15}{2.96} & \uentry{6.01}{5.92} & \bestentry{7.03}{6.74} \\
Optimal BST & \bestentry{74.02}{2.80} & \entry{71.56}{1.25} & \entry{73.07}{1.05} & \entry{72.73}{2.41} \\
Quickselect & \entry{2.24}{3.60} & \entry{1.79}{3.06} & \bestentry{4.07}{5.34} & \entry{1.67}{1.89} \\
Quicksort & \bestentry{61.45}{17.25} & \entry{54.03}{19.41} & \entry{59.20}{7.53} & \entry{55.34}{17.45} \\
Segments Intersect & \entry{93.85}{0.40} & \entry{93.97}{0.49} & \bestentry{94.09}{0.16} & \entry{93.66}{0.36} \\
Strongly Connected & \entry{23.03}{15.84} & \bestentry{32.01}{8.96} & \uentry{30.99}{7.06} & \uentry{30.39}{6.12} \\
Task Scheduling & \bestentry{81.71}{1.54} & \entry{81.12}{2.12} & \entry{80.61}{1.36} & \entry{80.05}{1.82} \\
Topological Sort  & \entry{62.55}{6.81} & \uentry{63.11}{10.34} & \uentry{67.81}{12.78} & \bestentry{74.81}{2.15} \\

        \bottomrule
    \end{tabular}
  \label{tab:results_mpnn}
\end{table}

\section{Attractors in Latent Spaces}

We now view latent representations through the lens of dynamical systems. In dynamical systems, a function governs how points move in ambient space over time. The connection with NARs is clear -- points are embeddings, the ambient space is the latent space, and steps of NAR are a discretisation of the function. We have already relied on this analogy implicitly, and have even been borrowing terminology such as \emph{trajectories} of points/embeddings. However, it is useful to make the connection explicit as this can reveal additional information about the latent spaces.

One important concept in the area of dynamical systems is that of attractors. Attractors are subsets of the latent space that are invariant under the update function and to which nearby embeddings are drawn to. Attractors can be points, cycles, or, as in the famous example of the Lorentz attractor, neither. 
Empirically (Fig.~\ref{fig:1}), we notice that the first few steps of execution create large movements in the latent space, and all steps from fifth onward seem to blend into each other. This suggests some form of convergence towards a single attractor. This can also be seen in terms of distances travelled at each time step, which quickly converge to zero. Note that different equivalent instances of the problem do not converge to the same point; we can view this as some form of partial (or multidimensional) attractor~\cite{Maass07}.

\begin{figure}[htbp]
    \centering
    \includegraphics[width=\textwidth]{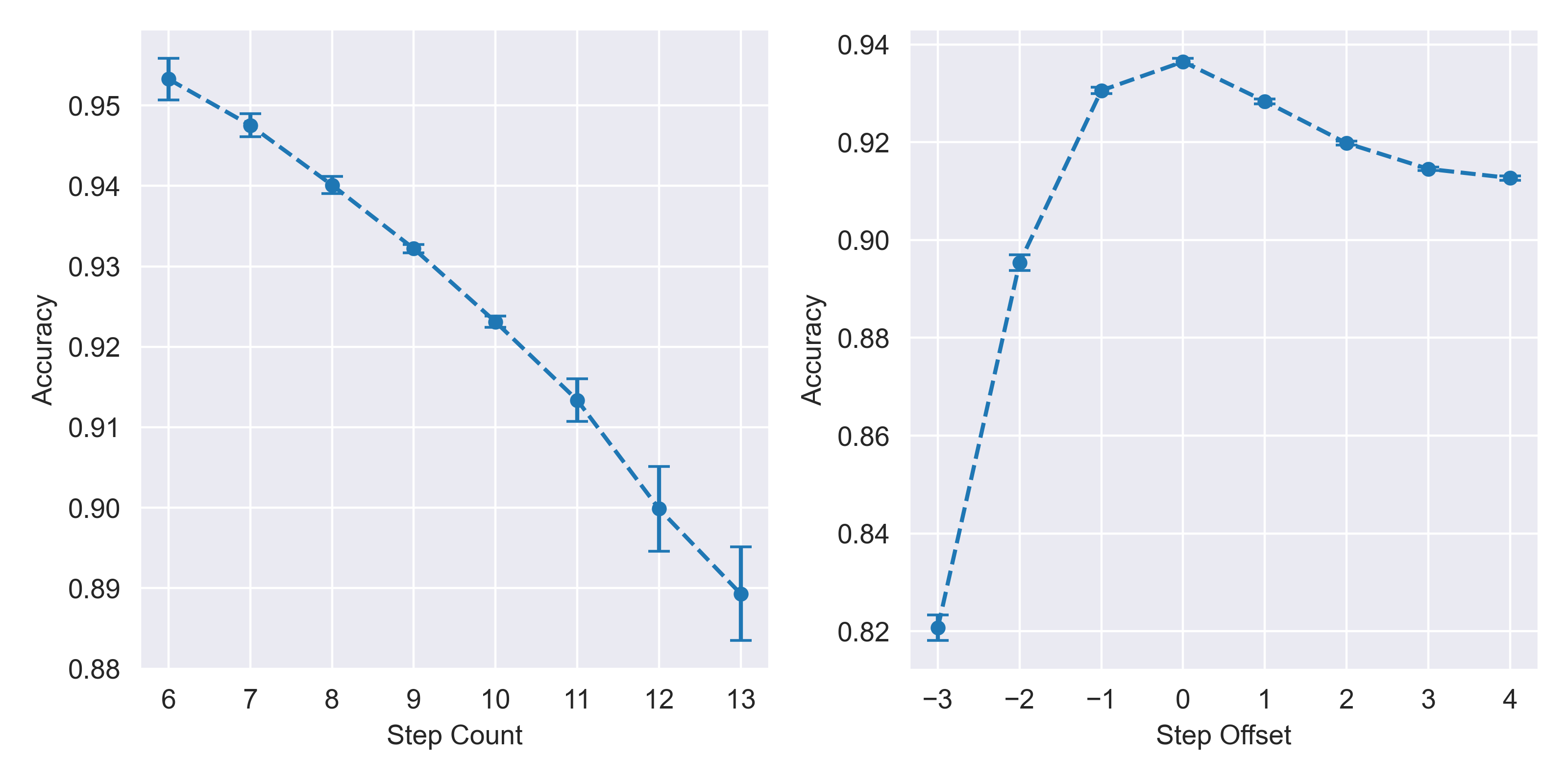}
    \caption{Left: Model accuracy versus the step count of Bellman-Ford execution. Graphs with lower step count perform better. Right: Model accuracy as their NAR execution is forcefully increased or decreased. Both reduce performance. Higher accuracy is better. }
    \label{fig:3}
\end{figure}

To test the properties of the attractor, we measure model accuracy as embeddings reach convergence. We separate the samples \emph{post hoc} based on the number of steps performed until completion. We discover that the accuracy is inversely proportional to the number of steps of execution of the algorithm (Fig.\ \ref{fig:3} left). Then, we forcefully modify the number of steps the model executes (Fig.\ \ref{fig:3} right). We observe that the performance suffers greatly when we stop earlier. This suggests that the model learns to be closely aligned with the ground truth Bellman-Ford algorithm. Specifically, it learns to always reason about new nodes on the same step that Bellman-Ford does. Stopping early, therefore, prevents it from completing its computation. Even more interestingly, performance also suffers when we run for longer than needed, though to a lesser degree. This is consistent with our previous observations, and shows that the attractor state is somewhat unstable under the model update.

\section{Ablations}

\subsection{Latent Spaces of Triplet-GMPNN}

\begin{figure}[htbp]
    \centering
    \includegraphics[width=\textwidth]{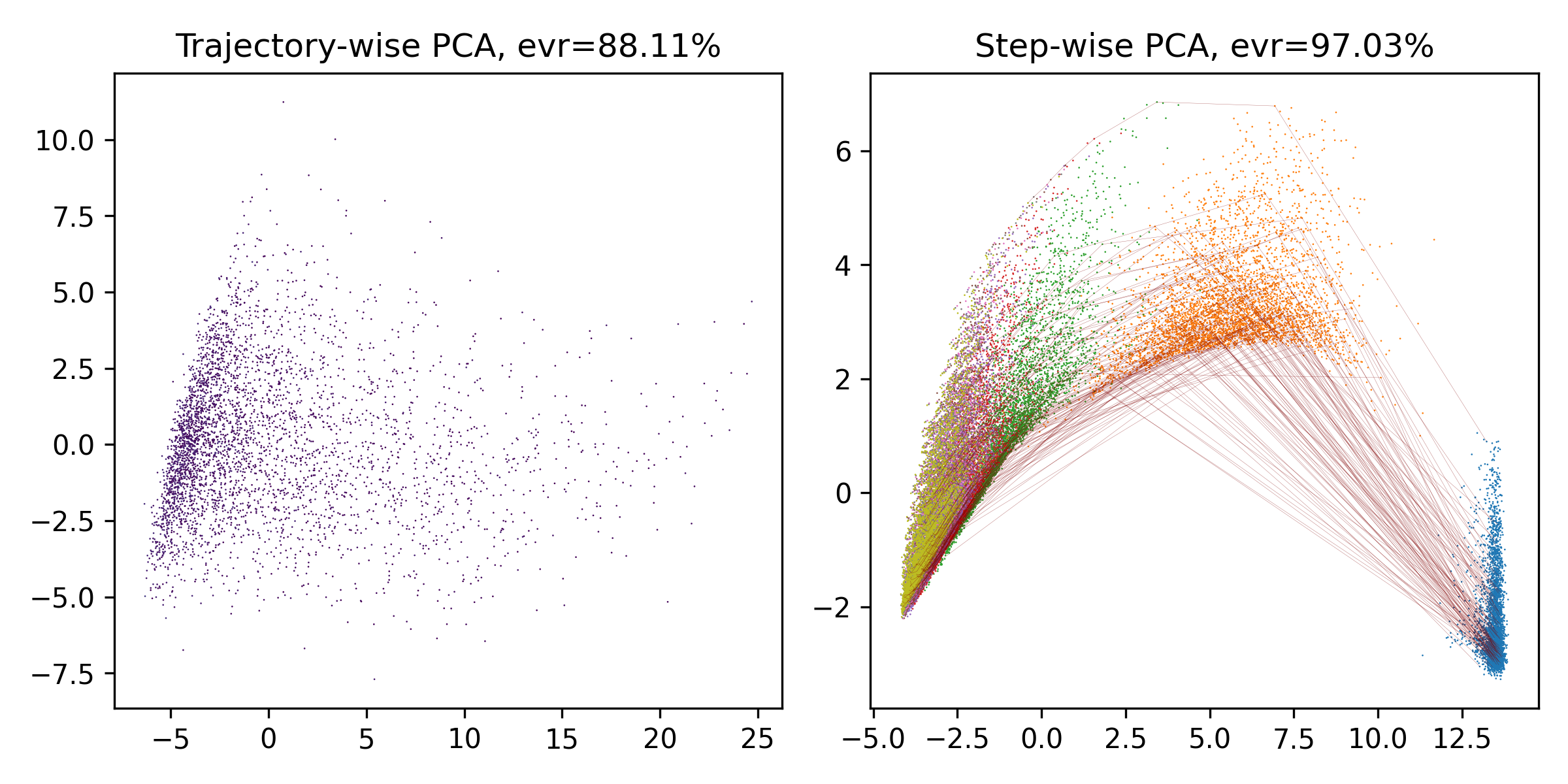}
    \caption{Latent space visualisation of Triplet-GMPNN.}
    \label{fig:tgmpnn_rw_2d}
\end{figure}

Our analysis of latent spaces was primarily focused on the LinearPGN. In Figure \ref{fig:tgmpnn_rw_2d}, we show the latent spaces for random ER graphs of Triplet-GMPNN. It is evident that the structure of latent spaces is similar to that of LinearPGN. Quantitatively, Triplet-GMPNN achieves slightly larger explained variance ratio, but qualitatively there is no substantial difference. 

In fact, the only aspect of our experiments that holds for LinearPGN but not for Triplet-GMPNN is the analysis of mispredicts in terms of value generalisation. Triplet-GMPNN achieves over $97\%$ accuracy on Bellman-Ford, and manages to generalise well to the out-of-distribution graphs. It learns to correctly handle large distances even without any changes to the pipeline, and we believe that this is precisely the reason for its improved performance.

Therefore, our work can be used to analyse other types of GNN processors, and is not limited to just one architecture.

\subsection{Choice of Node Aggregation}
\label{sec:aggro}

\begin{figure}[htbp]
    \centering
    \includegraphics[width=\textwidth]{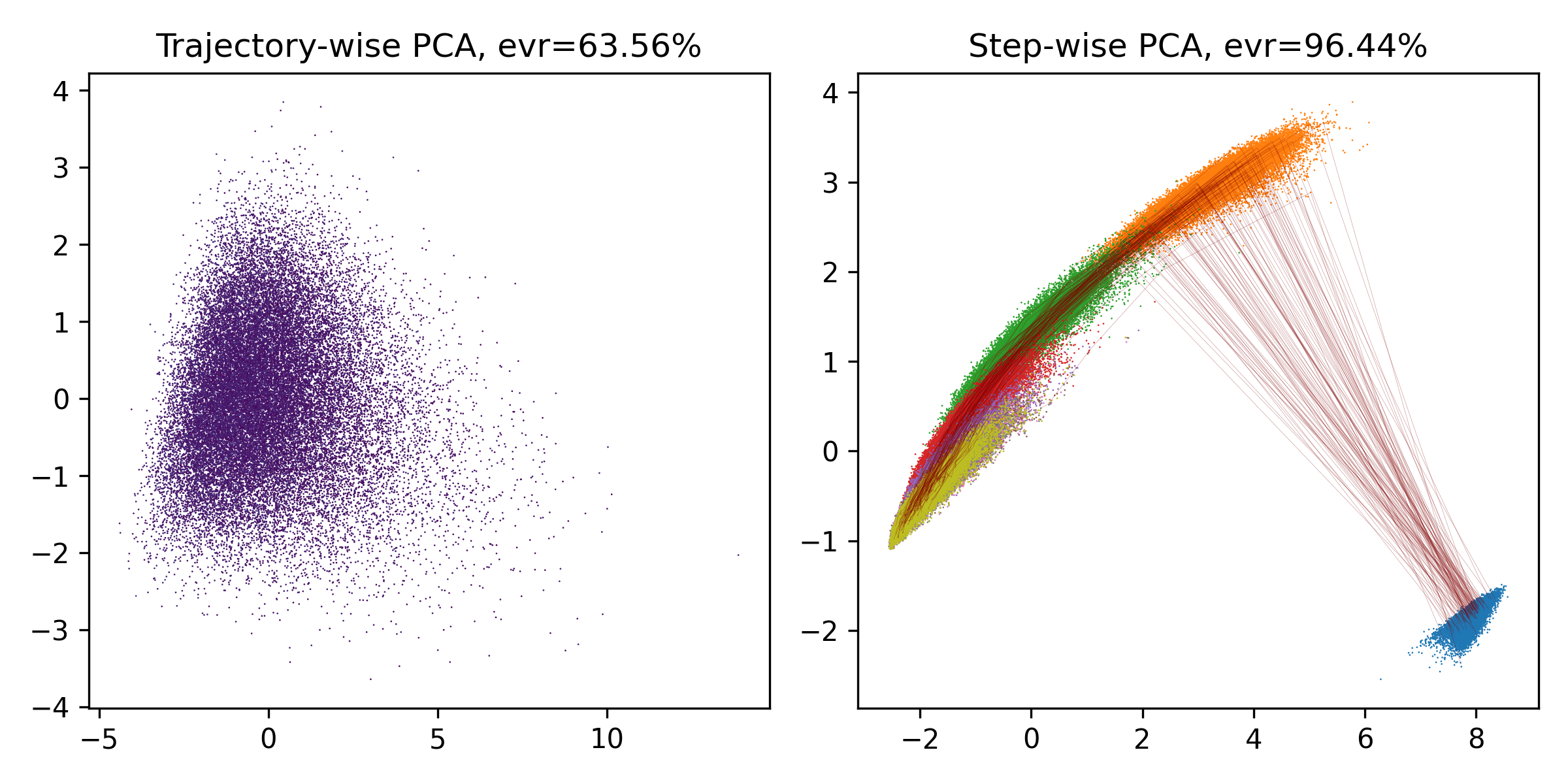}
    \caption{Visualisation of latent spaces with max aggregation.}
    \label{fig:refresh}
\end{figure}

In Section \ref{sec:trajs}, we introduced several approaches to using PCA in order to represent latent spaces. Common to all of the approaches was the need to reduce trajectory tensor's dimension from $4$ to $2$. In order to do so, we aggregated the trajectories for each node by selecting the largest values. As a refresher, visualisations of trajectory tensors reduced with $\max$ aggregation are shown on Figure \ref{fig:refresh}.

\begin{figure}[htbp]
    \centering
    \includegraphics[width=\textwidth]{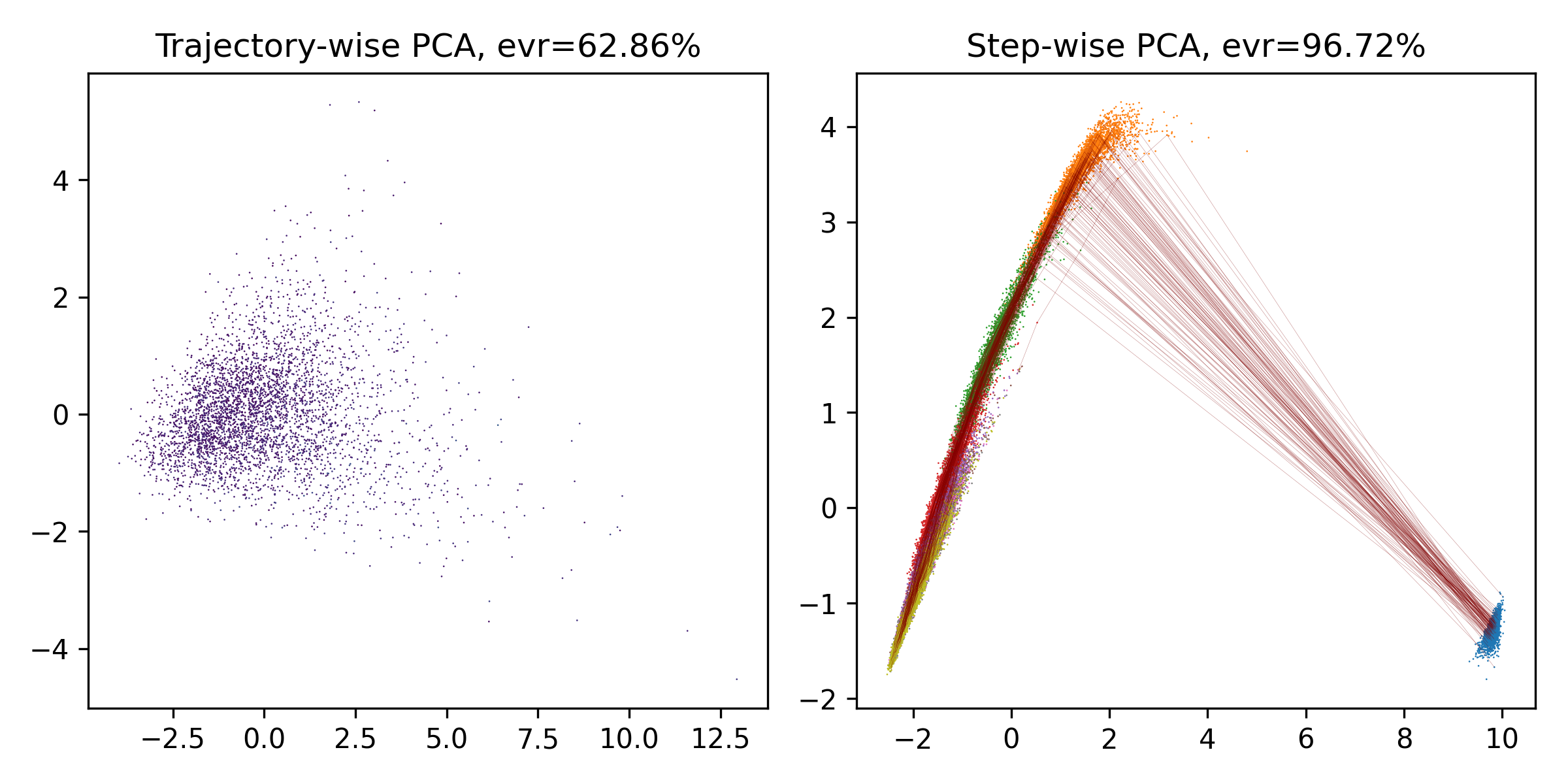}
    \caption{Visualisation of latent spaces with $\min$ aggregation.}
    \label{fig:min_aggro_2d}
\end{figure}

We could have done this differently, by using mean or $\min$ instead of $\max$. However, that would not have impacted our analysis. As an example, in Figure \ref{fig:min_aggro_2d} we recreate Figure \ref{fig:refresh} with $\min$ aggregation over nodes instead of with $\max$. There is no difference between the two approaches. Qualitatively, we observe the same visual structure; quantitatively, expected variance ratios for both trajectory-wise and step-wise PCA are nearly identical.

\begin{figure}[htbp]
    \centering
    \includegraphics[width=\textwidth]{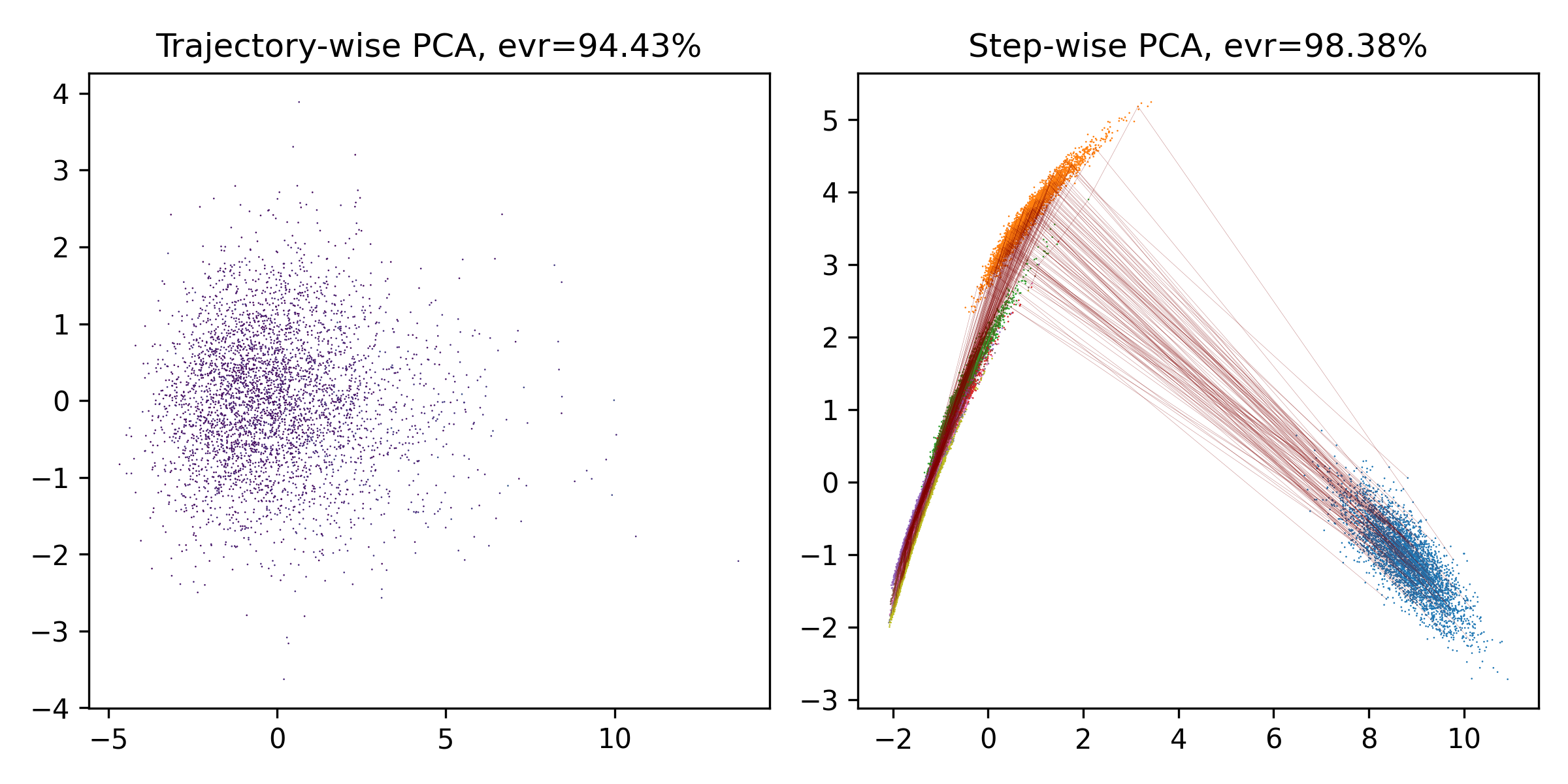}
    \caption{Visualisation of latent spaces with mean aggregation.}
    \label{fig:mean_aggro_2d}
\end{figure}

We visualise latent spaces with mean aggregation as well (Figure \ref{fig:mean_aggro_2d}). Again, there is no qualitative difference, and quantitatively, taking the mean increases the explained variance ratio of the trajectory-wise PCA from $63\%$ to $94\%$.

\subsection{Decay Strength}
\label{sec:decay}

The processor decay has a hyperparameter that determines its strength. We previously argued that only small amounts of decay are needed in order to sufficiently change the pipeline. 

Now, we look at different choices of the scaling factor and compare their performance. We also juxtapose decaying to zero, where we merely scale with the factor, against scaling differences to the mean.

From Table \ref{tab:decay} we observe that even decay with a factor of $0.9$, which corresponds to reducing embeddings by $10\%$ per step, is large enough to improve performance. Furthermore, both decaying towards zero, and decaying towards a mean that is calculated on a per-node basis, perform similarly. We conclude that mere presence of decay is enough to reshape the processor.

\begin{table}[htbp]
  \centering
    \begin{tabular}{lccc}
        \toprule
        \multicolumn{1}{c}{} & \multicolumn{3}{c}{Decay factor} \\
        \cmidrule(rl){2-4} 
                       Decay type & {$1$}  & {$0.9$} & {$0.5$} \\
        \cmidrule(r){1-1} \cmidrule(rl){2-4}
          To Zero     & \multirow{2}{*}{\entry{98.61}{0.10}} & \bestentry{98.76}{0.10} & \entry{98.64}{0.30} \\
          To Mean     &  & \entry{97.91}{044}     & \entry{98.63}{0.28} \\
        \bottomrule

    \end{tabular}
    \vspace{0.5em}
  \caption{Ablation of decay factor on Triplet-GMPNN on Bellman-Ford. Factor of $1$ means no decay, and smaller values increase decay strength. Standard deviation estimated with three runs. }
  \label{tab:decay}
\end{table}

\subsection{Softmax Temperature}

Our softmax aggregation also has a hyperparameter called temperature. Temperature controls how similar softmax is to hard $\max$, and how big an effect small values have on the result. Temperature is always non-negative. With temperature zero, softmax aggregation behaves exactly as $\max$. On the other hand, as temperature approaches $+\infty$, softmax aggregation behaves as mean. Thus, in order to preserve $\max$-like behaviour, temperature should be very small. But, in order to be differentiable through all elements, temperature should be larger than zero.

\begin{table}[htbp]
  \centering
    \begin{tabular}{lcccc}
        \toprule
          Temperature  & 0 & 0.01 & 0.1 & 1 \\
        \cmidrule(rl){1-5}
          Accuracy & \entry{98.68}{0.34} & \bestentry{98.80}{0.14} & \entry{97.99}{0.42} & \entry{98.29}{0.73}\\
        \bottomrule
    \end{tabular}
    \vspace{0.5em}
  \caption{Ablation of softmax temperature on Triplet-GMPNN on Bellman-Ford. Larger values of temperature increase impact of small arguments to softmax. Temperature of zero is equivalent to $\max$. Variance estimated with three runs.}
  \label{tab:softmax}
\end{table}

In Table \ref{tab:softmax}, we display the accuracy of Triplet-GMPNN when trained with different temperatures. We observe that softmax temperature of $0.01$ performs the best. Coincidentally, this is the temperature we used in our experiments above. We also note that large temperature also increases variance between results.

\section{Distance Distribution Changes}

We now turn to studying the effect of softmax aggregation and processor decay on the distribution of mispredicts.

\begin{figure}[htbp]
    \centering
    \includegraphics[width=\textwidth]{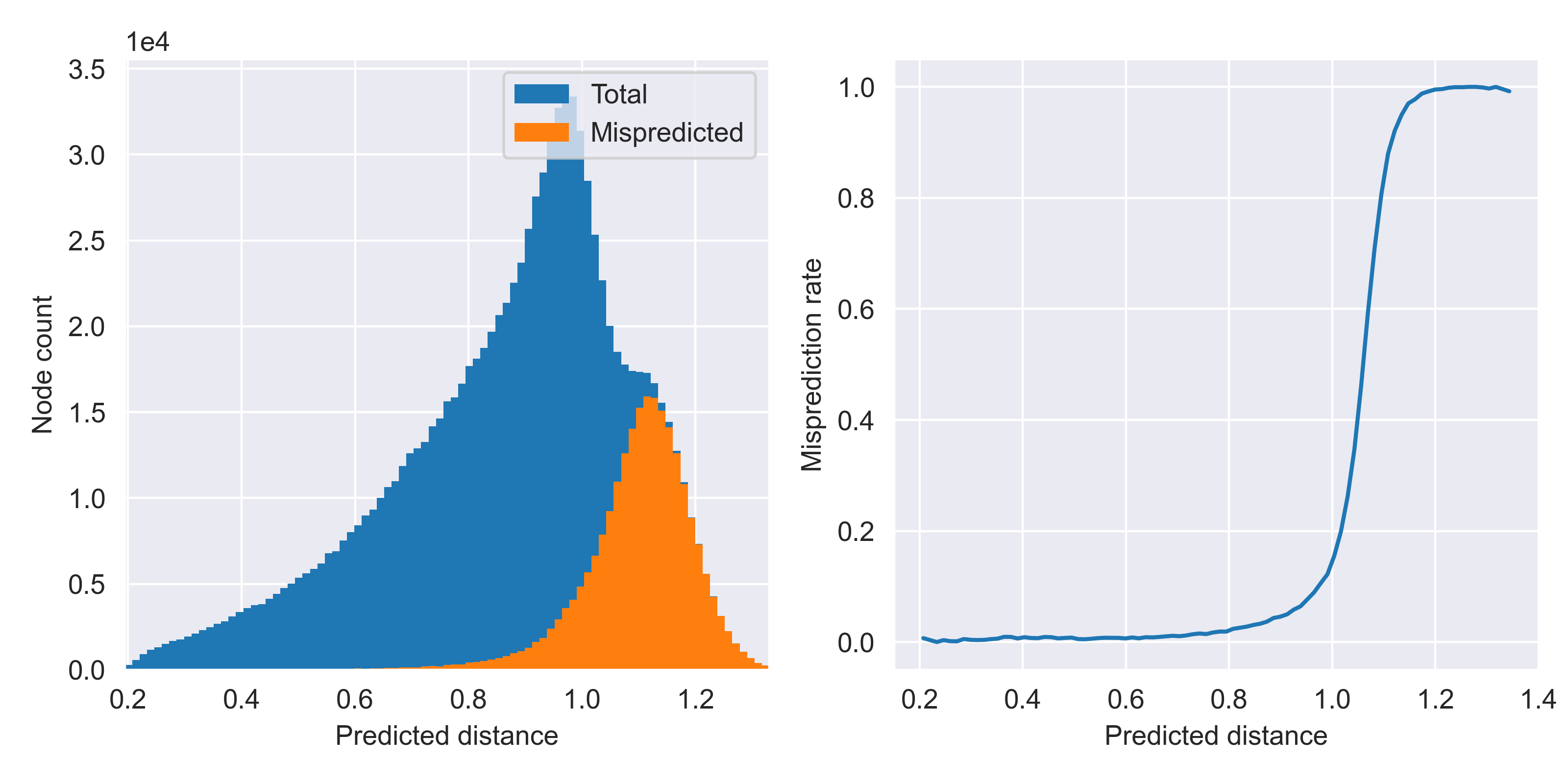}
    \caption{Distribution of predicted distances when NAR is trained with decay. Note that all mispredicts are clustered in a region to the right of $d=1$.}
    \label{fig:valgen_exp}
\end{figure}

\begin{figure}[htbp]
    \centering
    \includegraphics[width=\textwidth]{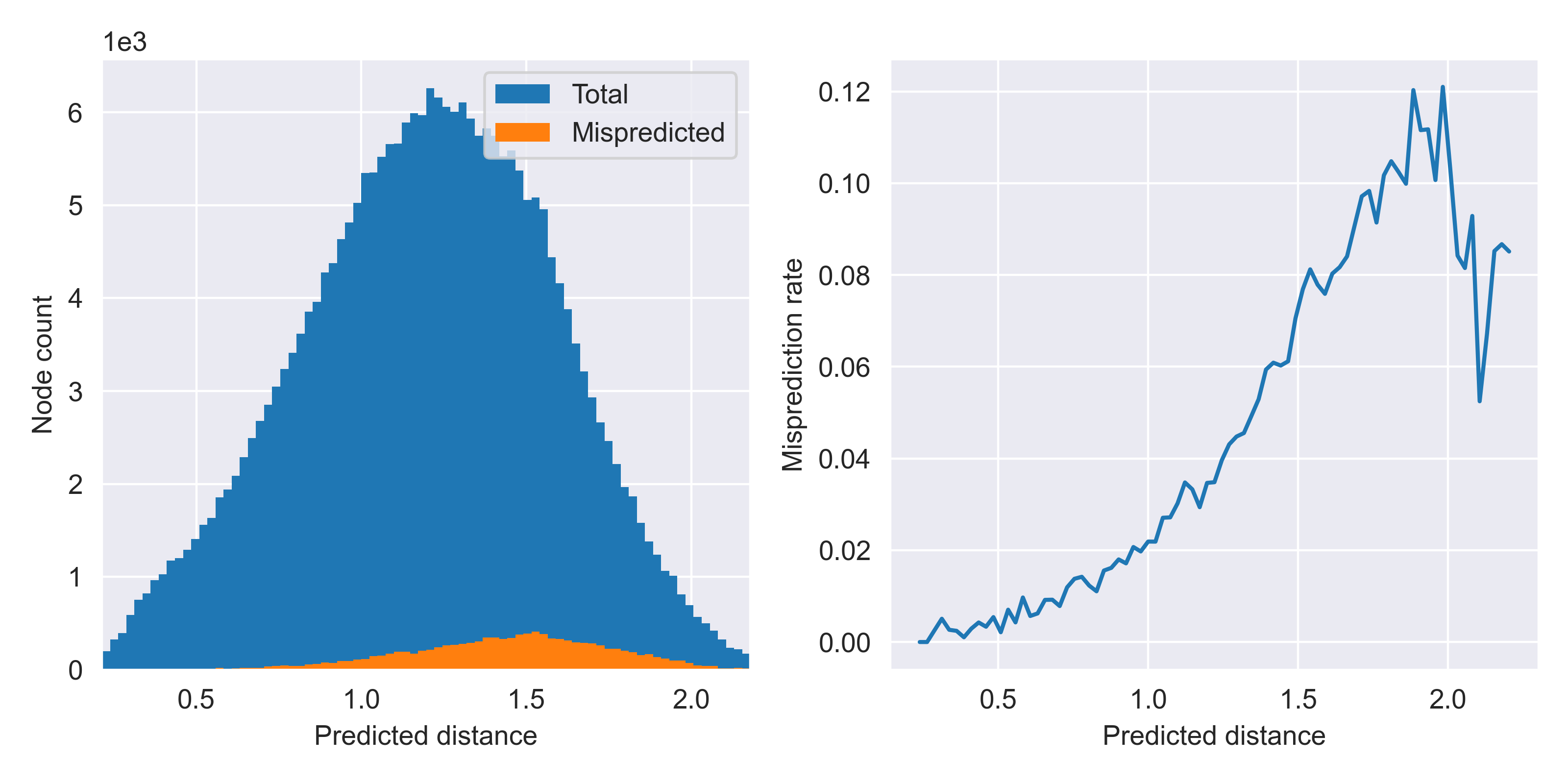}
    \caption{Distribution of predicted distances when NAR is trained with softmax aggregation. Note the scale on the right plot.}
    \label{fig:valgen_sft}
\end{figure}

Figures \ref{fig:valgen_exp} and \ref{fig:valgen_sft} show the distributions when NAR is trained with decay and softmax aggregation, respectively. In both cases, we observe that the distributions are drastically altered.

In the case of decay, we observe a clear separation between the correctly predicted and the mispredicted nodes. Almost all nodes with predicted distance under $1$ are predicted correctly, while almost all nodes with predicted distance above $1.2$ are mispredicted. With this approach, we can know with high confidence how trustworthy NAR predictions are.

\begin{figure}[htbp]
    \centering
    \includegraphics[width=\textwidth]{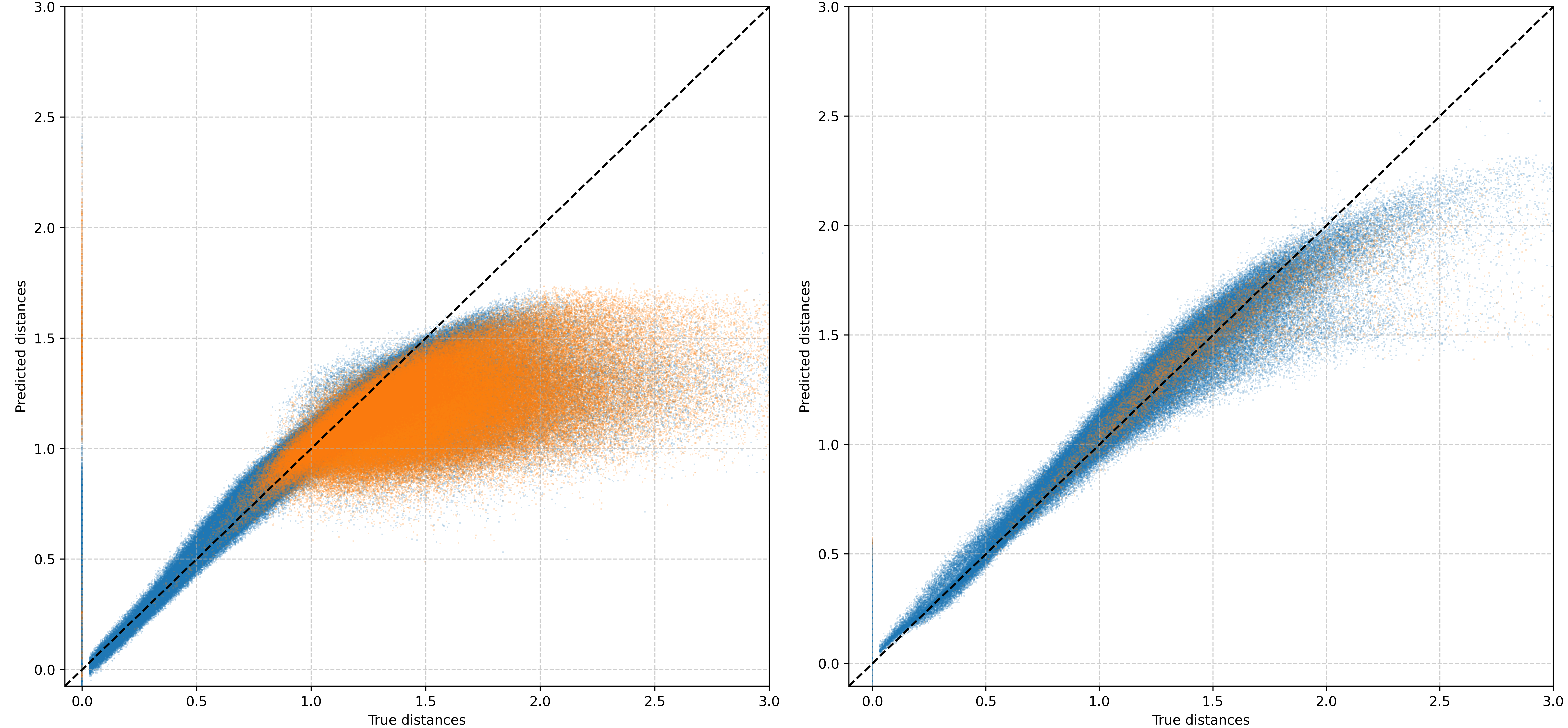}
    \caption{Left: Correlation between predicted and true distances for vanilla NAR. Right: Correlation for NAR with softmax aggregation. Correctly predicted nodes in blue, mispredicts in orange. }
    \label{fig:valgen_align}
\end{figure}

Softmax aggregation does not delineate between true and false predictions like decay does, but it greatly improves the overall performance. With softmax, NAR achieves impressive $96.5\%$ on the OOD $p=0.25$ dataset. In Figure \ref{fig:valgen_align} we plot the correlations between the true shortest distances and the ones predicted by NAR. This way, we can quantify how softmax improves the ability of the model to handle large unseen values.

\newpage 

\section{Bellman-Ford Prediction Visualisation}
\label{sec:example}

Here we visualise step-by-step predictions of vanilla Triplet-GMPNN NAR trained on Bellman-Ford algorithm, and how mispredicts can occur when comparing similar values.

Node 0 is source and is colored red, while others are gray. For each node, its id ($0$ to $9$) and predicted distance to source are overlaid on top.

To visualise predictions of node pointers $\pi$, we observe that they define a spanning tree with minimum distances to source. Therefore, we color each edge $uv$
\begin{itemize}
    \item \textcolor{blue}{Blue} if it belongs to a ground truth spanning tree ($\pi_\text{true}(v) = u$ or $\pi_\text{true}(u) = v$)
    \item \textcolor{green}{Green} if it belongs to a ground truth spanning tree \textbf{and} is correctly predicted by NAR ($\pi_\text{true}(v) = u$ or $\pi_\text{true}(u) = v$ \textbf{and} $\pi_\text{pred}(v) = u$ or $\pi_\text{pred}(u) = v$)
    \item \textbf{Black} if it does not belong to a ground truth spanning tree ($\pi_\text{true}(v) \neq u$ and $\pi_\text{true}(u) \neq v$)
    \item \textcolor{red}{Red} if it does not belong to a ground truth spanning tree \textbf{but} is mispredicted  as one ($\pi_\text{true}(v) \neq u$ and $\pi_\text{true}(u) \neq v$ \textbf{but} $\pi_\text{pred}(v) = u$ or $\pi_\text{pred}(u) = v$)
\end{itemize}

The graph is undirected.

\begin{figure}[htbp]
    \centering
    \includegraphics[width=\textwidth]{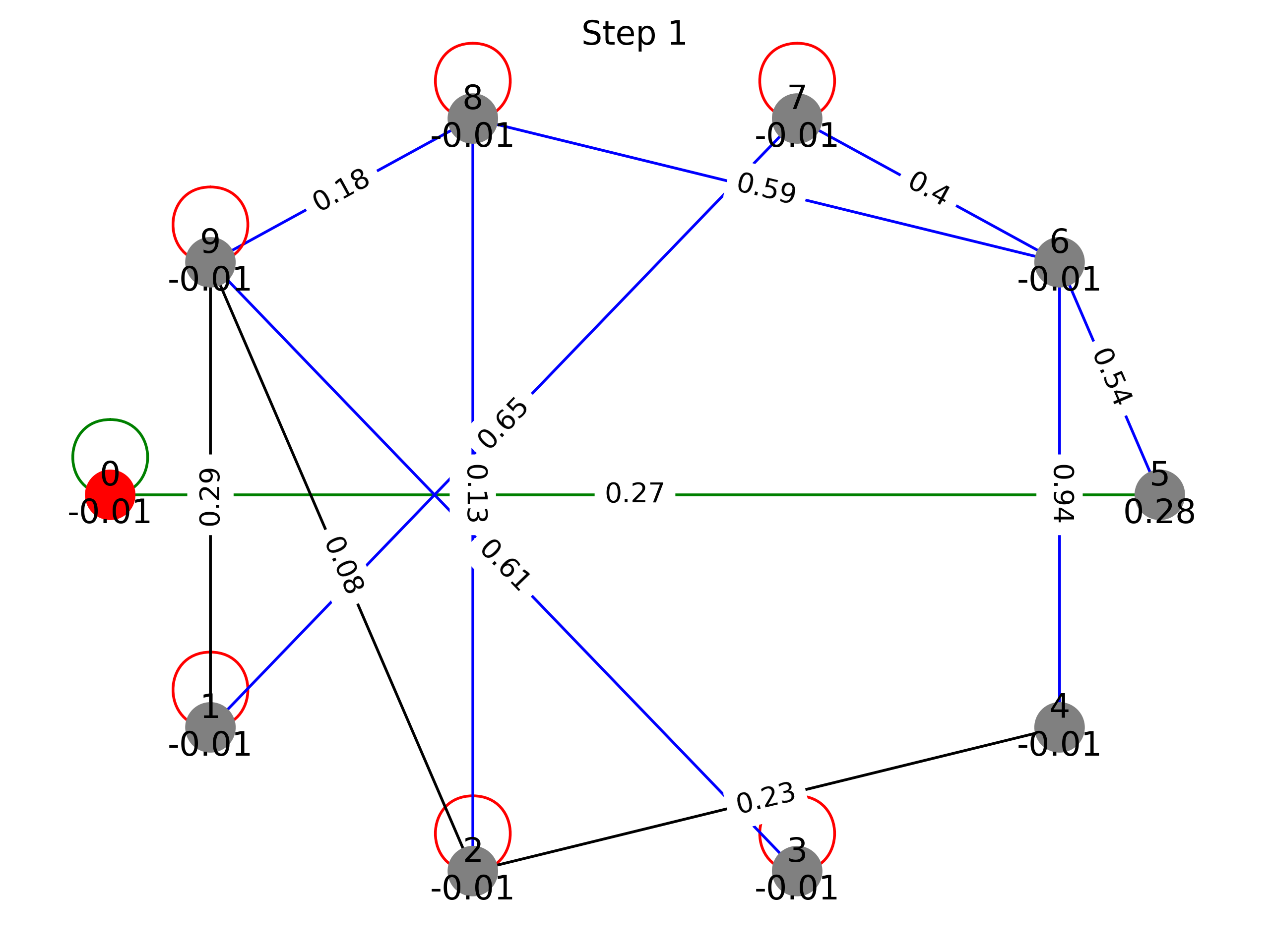}
\end{figure}

After step 1, node 5 is correctly directed to source. None of the other nodes are reachable yet.

After step 2, node 6 is also correctly directed to source, and after step 3, so are nodes 4, 7 and 8.

\begin{figure}[htbp]
    \centering
    \includegraphics[width=\textwidth]{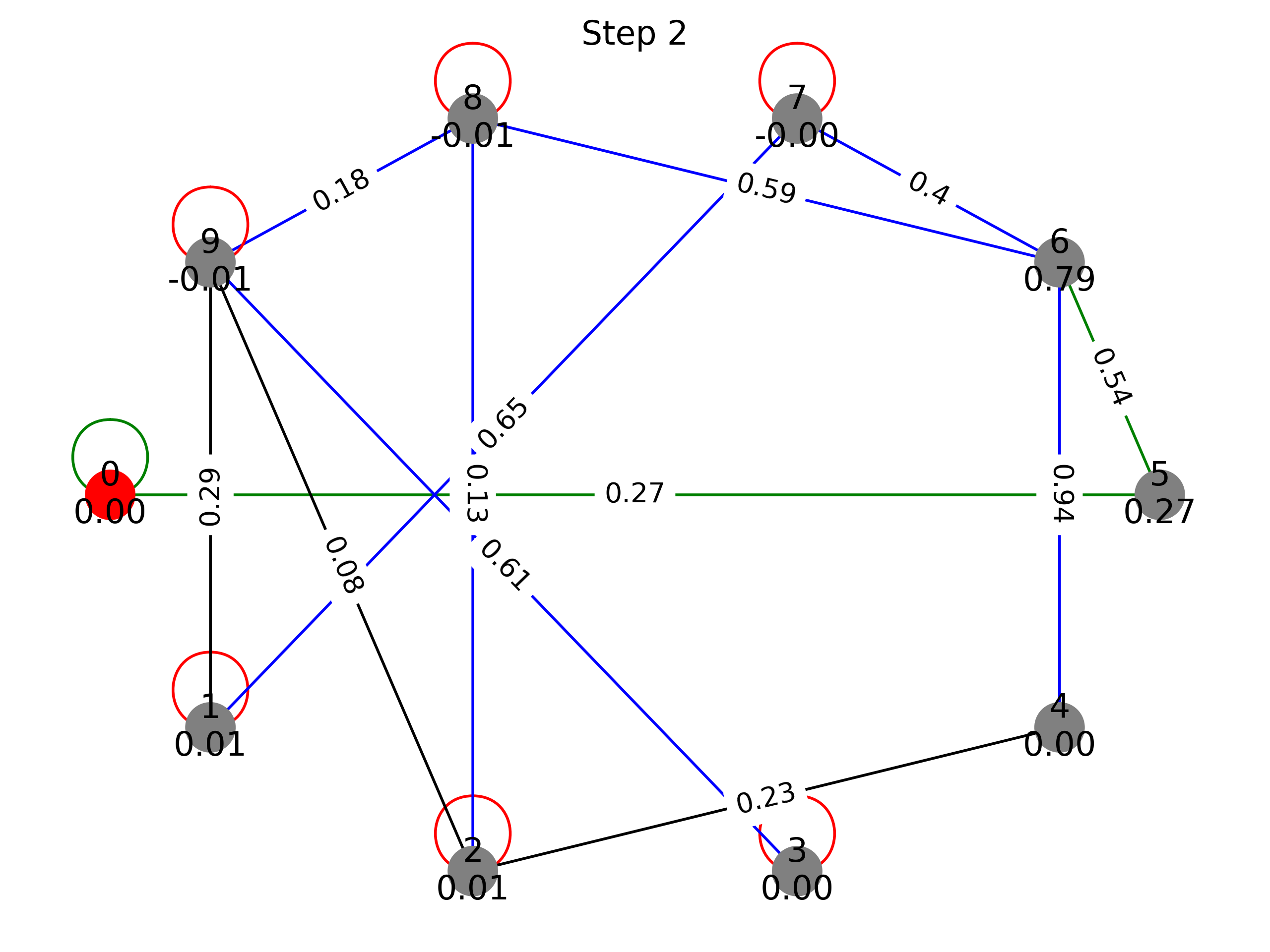}
\end{figure}

\begin{figure}[htbp]
    \centering
    \includegraphics[width=\textwidth]{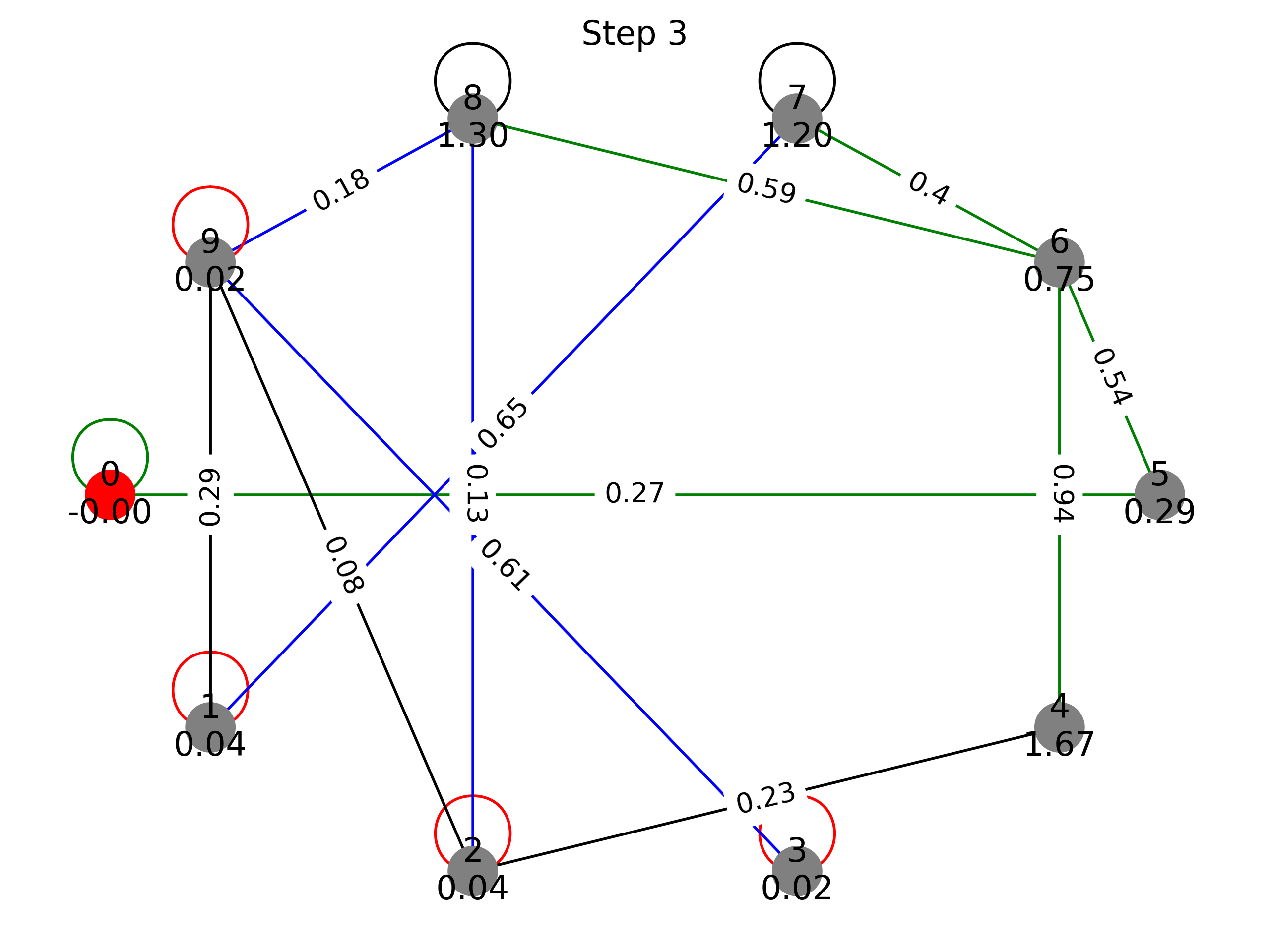}
\end{figure}

\begin{figure}[htbp]
    \centering
    \includegraphics[width=\textwidth]{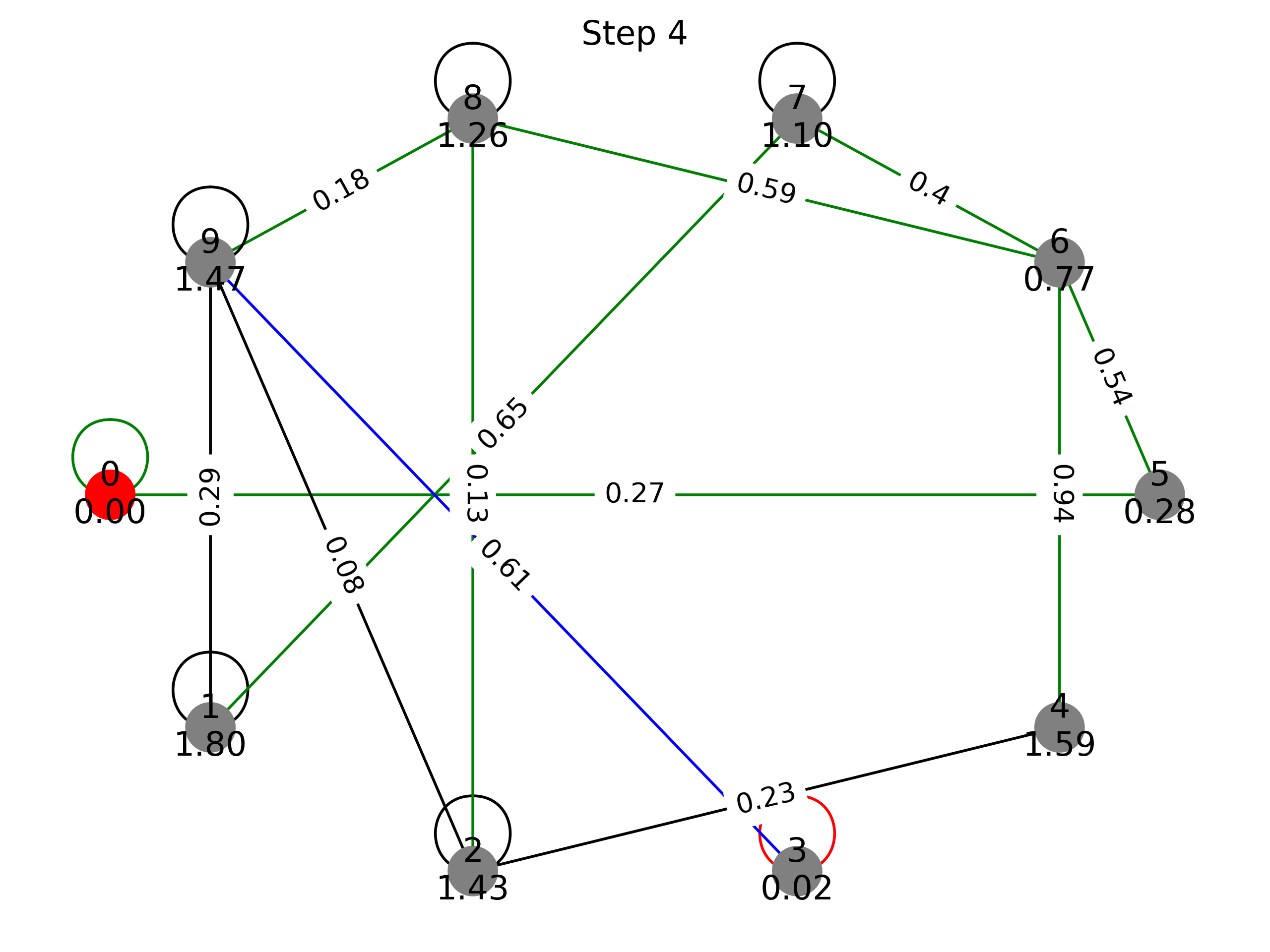}
\end{figure}

With step 4 complete, all discovered nodes are predicted correctly. One more step needs to be made to reach node 3.

\begin{figure}[htbp]
    \centering
    \includegraphics[width=\textwidth]{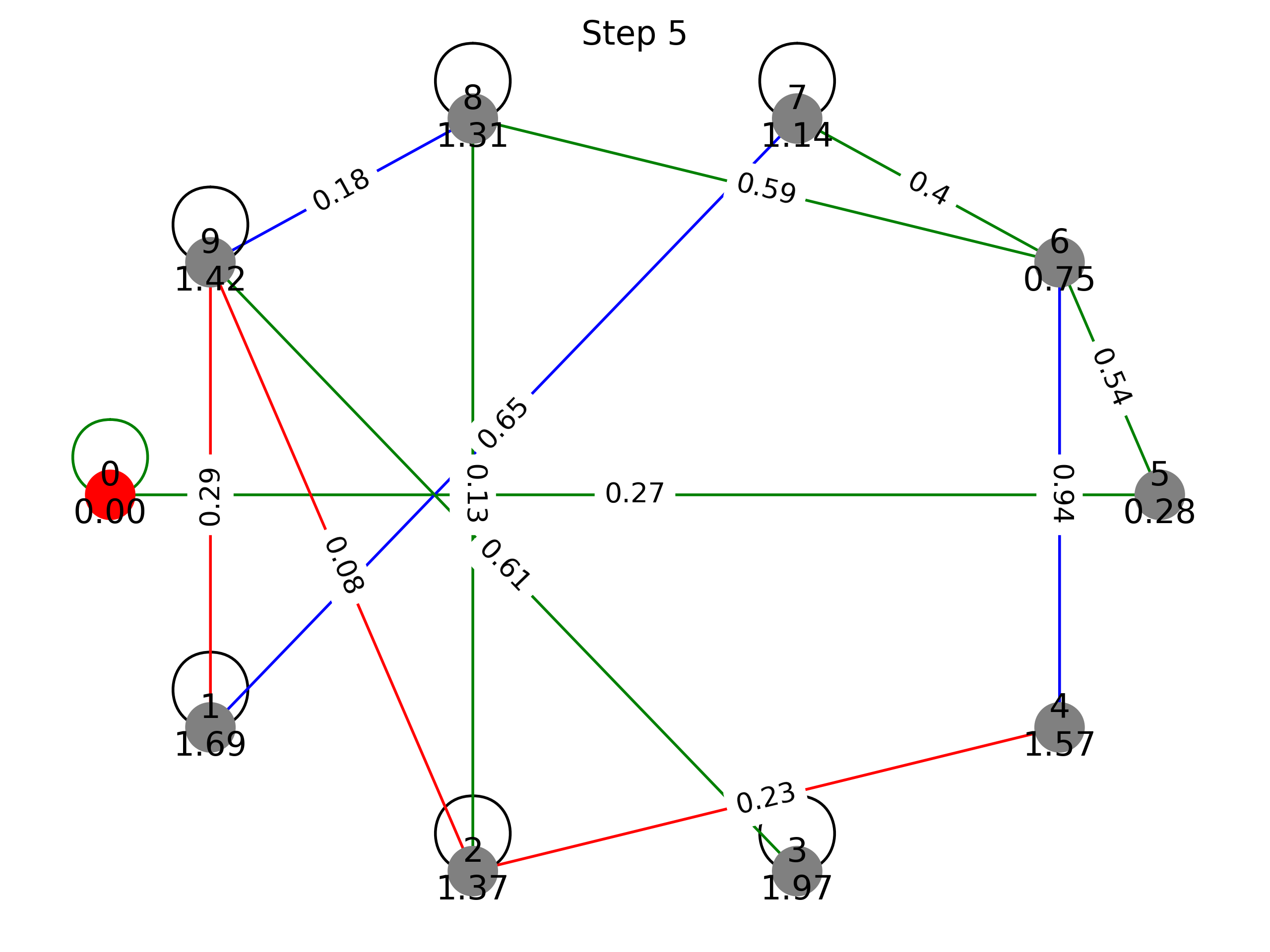}
\end{figure}

In this final step, neural network suddenly makes three mispredictions. It sets $\pi(4)=2$ from previous 6, $\pi(9)=2$ from previous 8, and $\pi(1)=9$ from previous 7. Crucially, in all of those cases, predicted distances to source between these pairs of paths are similar ($1.69$ vs $1.6$, $1.41$ vs $1.45$, and $1.79$ vs $1.71$ respectively). 

This failure mode is present among all Bellman-Ford mispredicts we observed, and is the argument for using softmax aggregation outlined in Section 6.

\newpage

\section{Latent Space Trajectories - Sequential View}
\label{sec:sideways}

\begin{figure}[htbp]
    \centering
    \includegraphics[width=\textwidth, trim={12cm 3.5cm 10.8cm 4cm},clip]{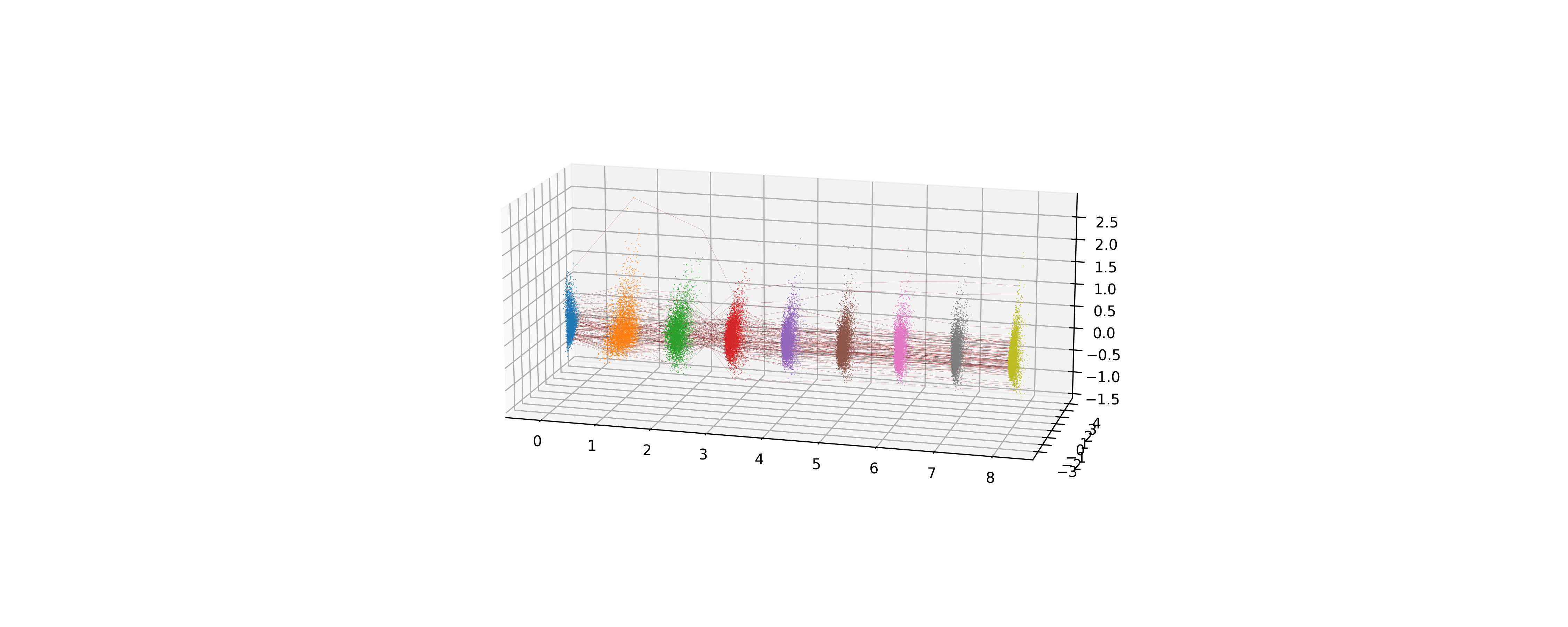}
    \caption{Sequential view of embedding trajectories in Figure \ref{fig:1} left.}
\end{figure}

\begin{figure}[htbp]
    \centering
    \includegraphics[width=\textwidth, trim={12cm 3.5cm 10.8cm 4cm},clip]{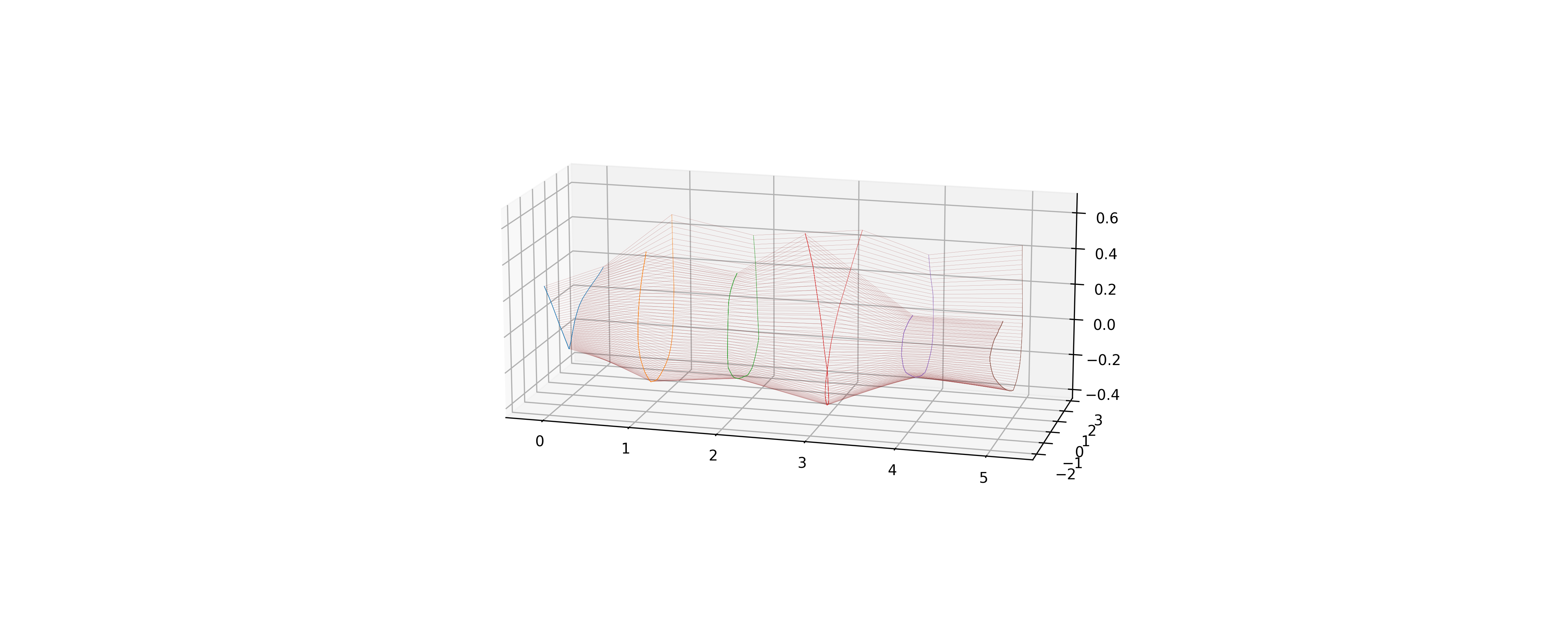}
    \caption{Sequential view of embedding trajectories in Figure \ref{fig:1} middle.}
\end{figure}

On these figures we visualise the embeddings by ordering the execution steps from left to right and tracing the trajectories in red. We see that the first few steps greatly change the step embeddings,
but that trajectories switch to being near-parallel as they approach the attractor.

\section{3D Visualisations}

Here, we display high-resolution visualisations of latent spaces, both trajectory-wise (left) and step-wise (right).

\begin{figure}[htbp]
    \centering
    \includegraphics[width=\textwidth]{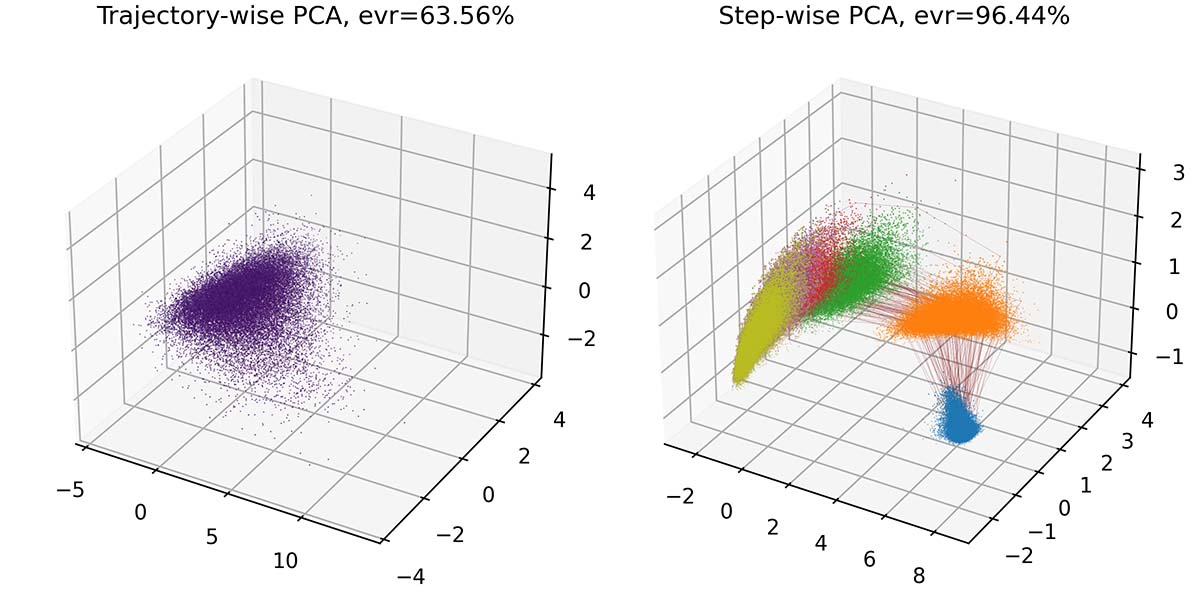}
    \caption{3D visualisation of random ER graphs.}
\end{figure}

\begin{figure}[htbp]
    \centering
    \includegraphics[width=\textwidth]{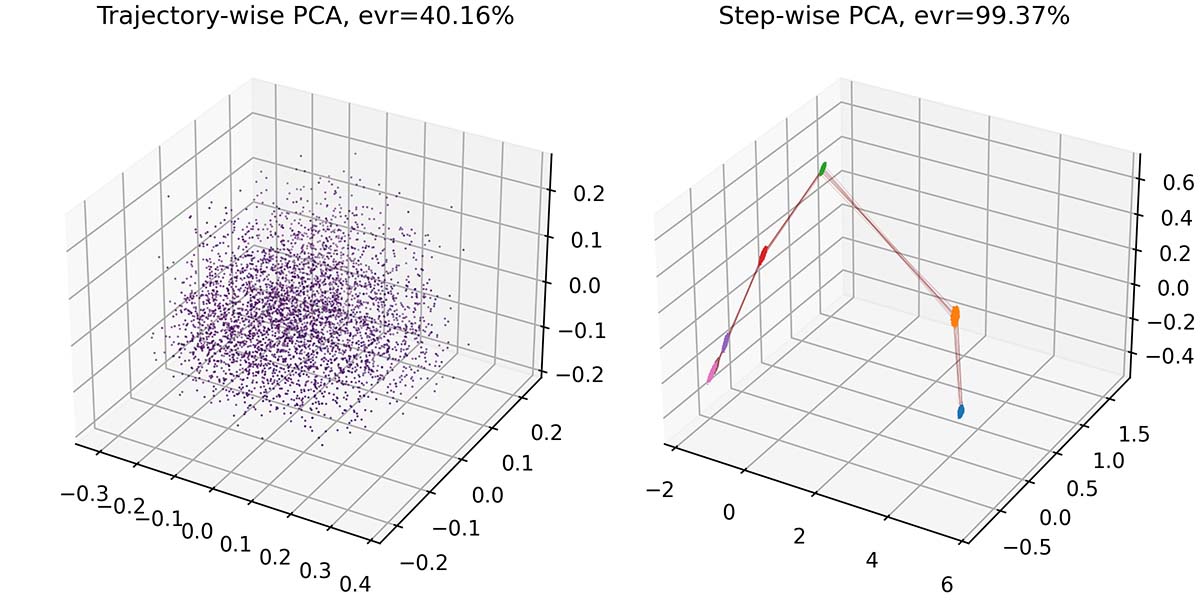}
    \caption{3D visualisation of permutation symmetric graphs.}
\end{figure}

\begin{figure}[htbp]
    \centering
    \includegraphics[width=\textwidth]{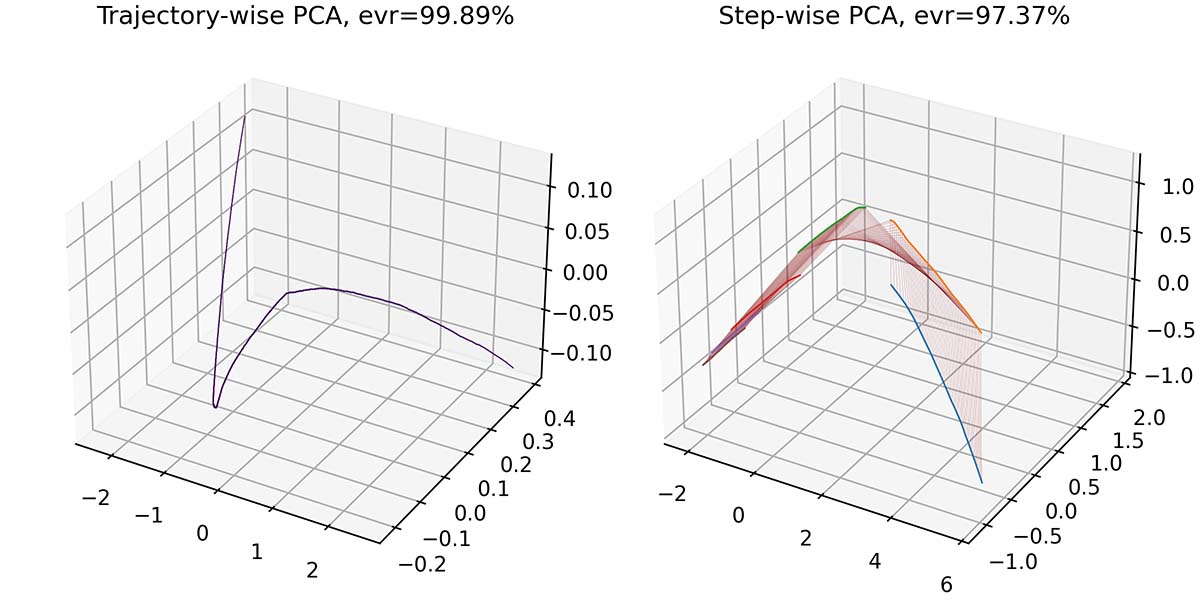}
    \caption{3D visualisation of scale symmetric graphs.}
\end{figure}

\begin{figure}[htbp]
    \centering
    \includegraphics[width=\textwidth]{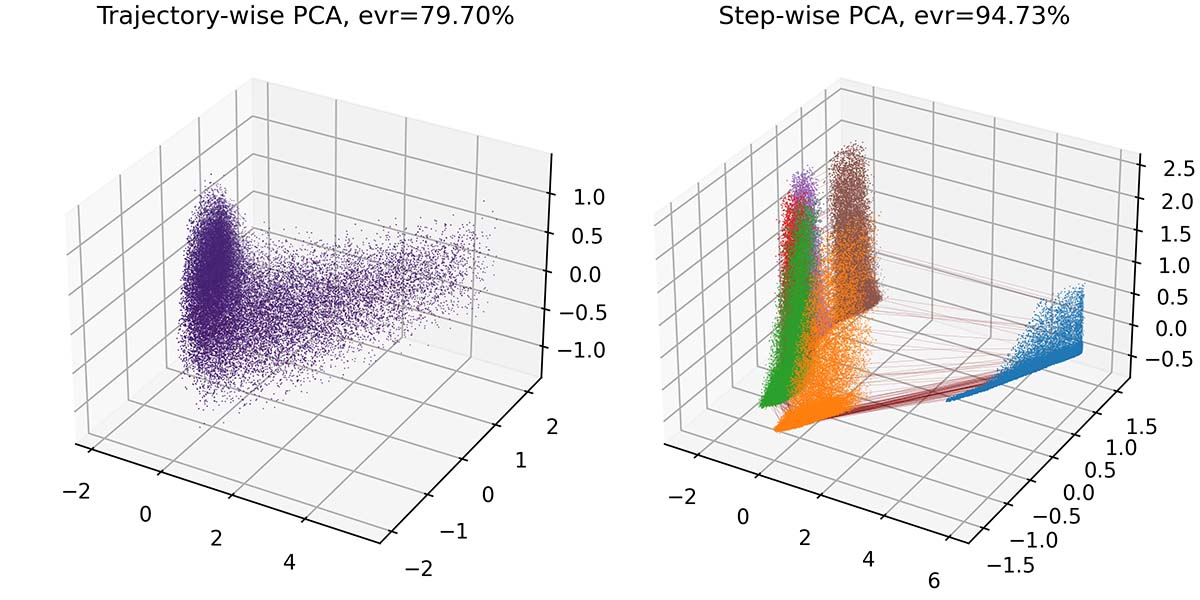}
    \caption{3D visualisation of reweighting symmetric graphs.}
\end{figure}

\begin{figure}[htbp]
    \centering
    \includegraphics[width=\textwidth]{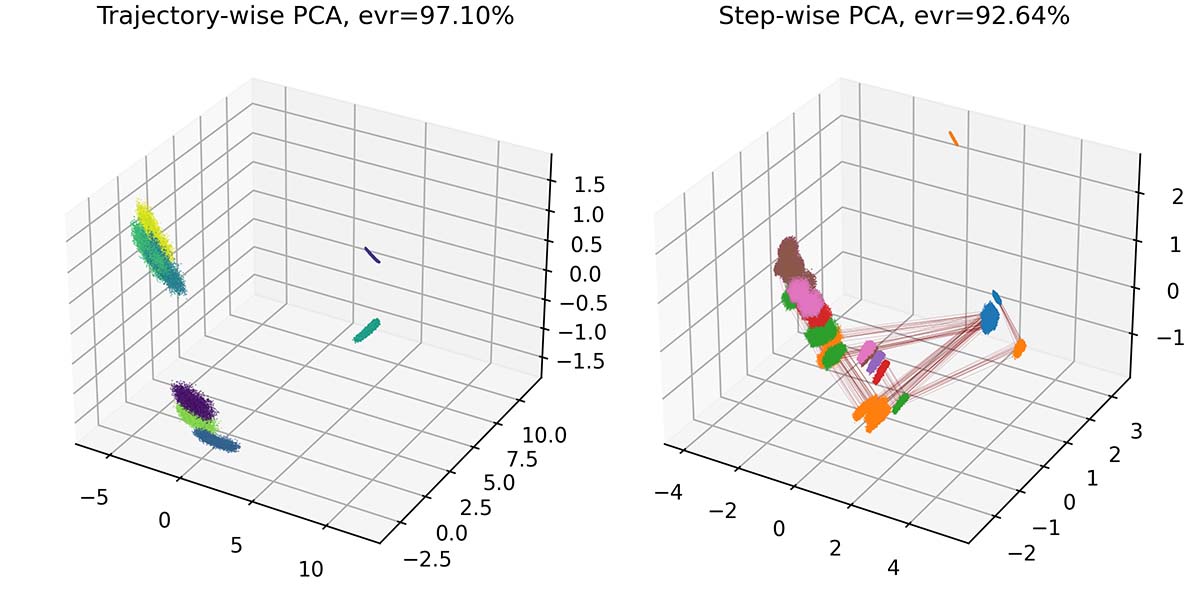}
    \caption{3D visualisation of reweighting symmetric graphs, with eight different clusters.}
\end{figure}

\end{document}